\documentclass[runningheads]{llncs}
\usepackage{graphicx}
\usepackage{amsmath,amssymb} 
\usepackage{color}
\usepackage[width=122mm,left=12mm,paperwidth=146mm,height=193mm,top=12mm,paperheight=217mm]{geometry}

\newcommand{\ra}[1]{\renewcommand{\arraystretch}{#1}}
\usepackage{gensymb}
\usepackage{amsfonts}
\usepackage{setspace}
\usepackage{booktabs}
\usepackage{footnote}
\makesavenoteenv{table}
\makesavenoteenv{tabular}
\RequirePackage{pdfpages}
\usepackage{multirow}
\newcommand{\minitab}[2][l]{\begin{tabular}{#1}#2\end{tabular}} 

\usepackage{caption}
\usepackage[labelformat=simple]{subcaption}
\usepackage{cite}
\usepackage{hyperref}

\newcommand\blfootnote[1]{%
  \begingroup
  \renewcommand\thefootnote{}\footnote{#1}%
  \addtocounter{footnote}{-1}%
  \endgroup
}

\begin{document}

\pagestyle{headings}
\mainmatter

\title{Nested Invariance Pooling and RBM Hashing for Image Instance Retrieval} 
\author{Olivier Mor\`{e}re{$^{1,2}$}, 
Jie Lin{$^{1}$}, 
Antoine Veillard{$^{2}$},\\ 
Vijay Chandrasekhar{$^{1}$},
Tomaso Poggio{$^{3}$}
}
\institute{Institute for Infocomm Research, A*STAR, Singapore{$^{1}$}\\Universit\'e Pierre et Marie Curie, Paris, France{$^{2}$}\\CBMM, LCSL, IIT and MIT, Cambridge, USA{$^{3}$}}
\maketitle
\blfootnote{Authors contributed equally to this work.}

\begin{abstract}

The goal of this work is the computation of very compact binary hashes for image instance retrieval.
Our approach has two novel contributions.
The first one is Nested Invariance Pooling (NIP), a method inspired from \emph{i-theory}, a mathematical theory for computing group invariant transformations with feed-forward neural networks.
NIP is able to produce compact and well-performing descriptors with visual representations extracted from convolutional neural networks.
We specifically incorporate scale, translation and rotation invariances but the scheme can be extended to any arbitrary sets of transformations.
We also show that using moments of increasing order throughout nesting is important.
The NIP descriptors are then hashed to the target code size (32-256 bits) with a Restricted Boltzmann Machine with a novel batch-level regularization scheme specifically designed for the purpose of hashing (RBMH). 
A thorough empirical evaluation with state-of-the-art shows that the results obtained both with the NIP descriptors and the NIP+RBMH hashes are consistently outstanding across a wide range of datasets.


\end{abstract}

\section{Introduction}
\label{sec:intro}

Small binary image representations such as 64-bit hashes are a definite must for fast image instance retrieval.
Compact hashes provide more than enough capacity for any practical purposes, including internet-scale problems.
In addition, a compact hash is directly addressable in RAM and enables fast matching using ultra-fast Hamming distances.

State-of-the-art global image descriptors such as Fisher Vectors (FV)~\cite{Perronnin_CVPR_10}, Vector of Locally Aggregated Descriptors (VLAD)~\cite{PQFisher} and 
Convolutional Neural Network (CNN) features \cite{AlexNet,Yandex} allow for robust image matching.
However, the dimensionality of such descriptors is typically very high: 
4096 to 65536 floating point numbers for FVs\cite{Perronnin_CVPR_10} and 4096 for CNNs \cite{AlexNet}.
Bringing such high-dimensional representations down to compact hashes is a considerable challenge.

Deep learning has achieved remarkable success in many visual tasks such as image classification~\cite{AlexNet,Simonyan2014}, 
image retrieval~\cite{Yandex}, face recognition~\cite{deepface,deepid} and pose estimation~\cite{deeppose}. 
Here, we propose a deep learning framework for binary hashing that generates extremely compact, yet discriminative descriptors.  
A series of nested pooling layers introduced in the pipeline, 
provide higher invariance and robustness to common transformations like rotation and scale.
A RBM layer for hashing is introduced at the end of the pipeline to map real-valued data to binary hashes. 
The proposed deep learning pipeline generates hashes that consistently and significantly outperform other state-of-the-art methods 
at code size from 256 down to very small sizes like 32 bits, on several popular benchmarks.

\section{Background and Related Work}

Our image instance retrieval pipeline starts with the computation of high-dimensional vectors referred to as {\it global descriptors}, 
followed by a hashing step to obtain compact representations.
Here, we review the state-of-the-art in {\it global descriptors} and hashing methods.

{\textbf{Global Descriptors}.}
State-of-the-art {\it global descriptors} for image instance retrieval are based on 
either FV~\cite{Perronnin_CVPR_10}/VLAD~\cite{PQFisher} or Convolutional Neural Networks (CNN)~\cite{Yandex}.
Several variants of FV/VLAD~\cite{TEDA,confused,REVV1,SFCV} have been proposed, since it was first proposed for instance retrieval~\cite{Perronnin_CVPR_10}.
In recent work, CNN descriptors have begun being applied to the computation of global descriptors for image retrieval~\cite{Razavian2014,Yandex,NewPaperA,sharif2015baseline,babenko2,ToliasSJ15}.

Razavian et al.~\cite{Razavian2014} evaluate the performance of CNN activations from fully connected layer 
on a wide range of tasks including instance retrieval, and show initial promising results.
After that, Babenko et al.~\cite{Yandex} show that a pre-trained CNN can be fine tuned with domain specific data (objects, scenes, etc.) 
to improve retrieval performance on relevant datasets.
In~\cite{NewPaperA}, the authors propose extracting activations of fully connected layer from multiple regions sampled in an image, 
followed by aggregating VLAD descriptors on these local CNN activations.
While this results in highly performant descriptors, the starting representations are orders of magnitude larger than descriptors proposed in this work.
\cite{Razavian15,sharif2015baseline} show that spatial max pooling of intermediate maps is an effective representation and higher performance can be achieved compared to using the fully connected layers.
Babenko et al~\cite{babenko2} in their very recent work, show that sum-pooling of intermediate feature maps performs better than max-pooling, when the image representation is whitened.
Note that the approach in~\cite{babenko2} provide limited invariance to translation, but not to scale or rotation.
Another very recent work~\cite{ToliasSJ15} proposes pooling across regional bounding boxes in the image, similar to the popular R-CNN approach~\cite{girshick2014rcnn} used for object detection.

\begin{figure}
\centering{
	\begin{tabular}{@{}c@{}@{}c@{}}
		\includegraphics[width=0.45\textwidth]{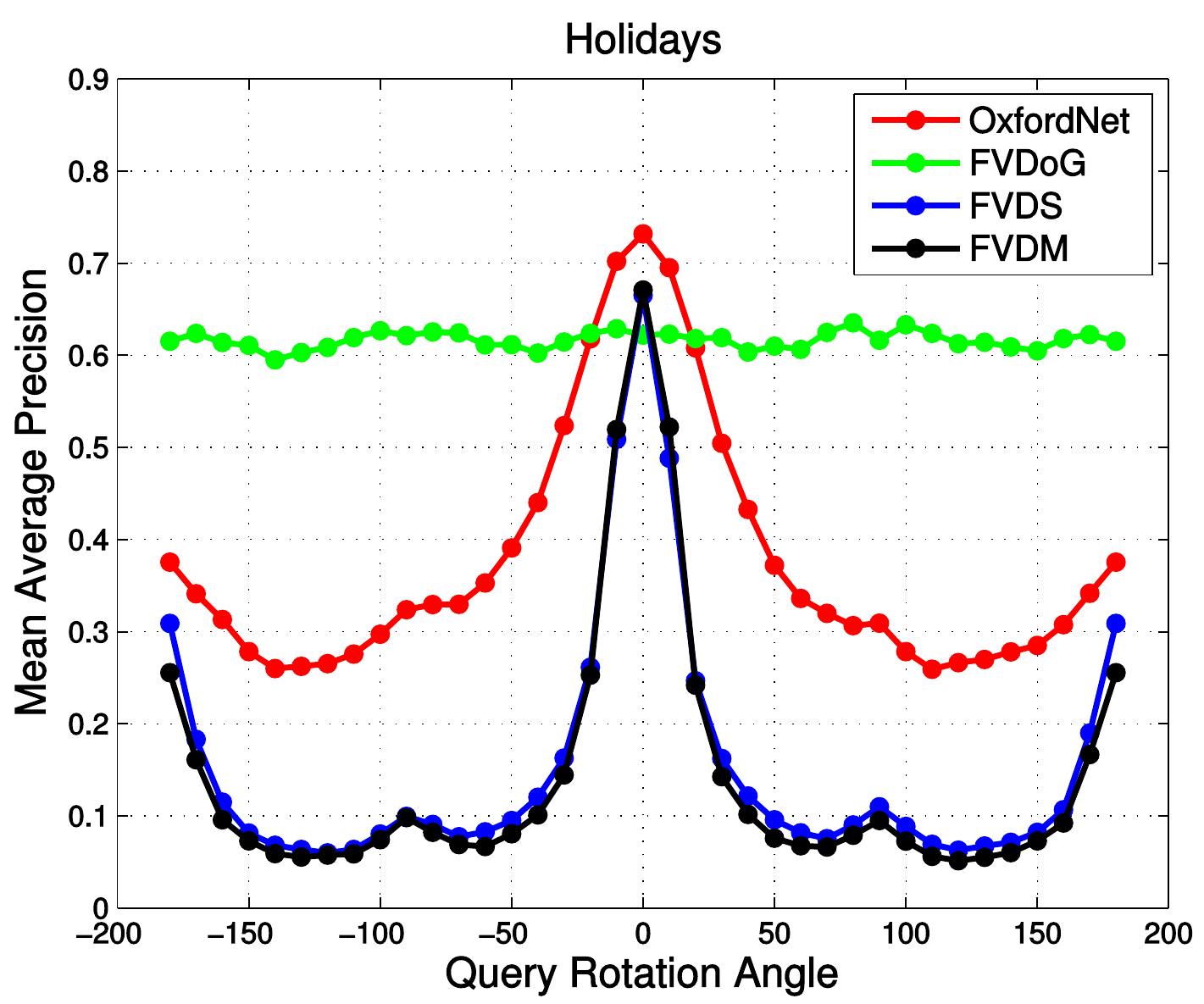} &
		\includegraphics[width=0.45\textwidth]{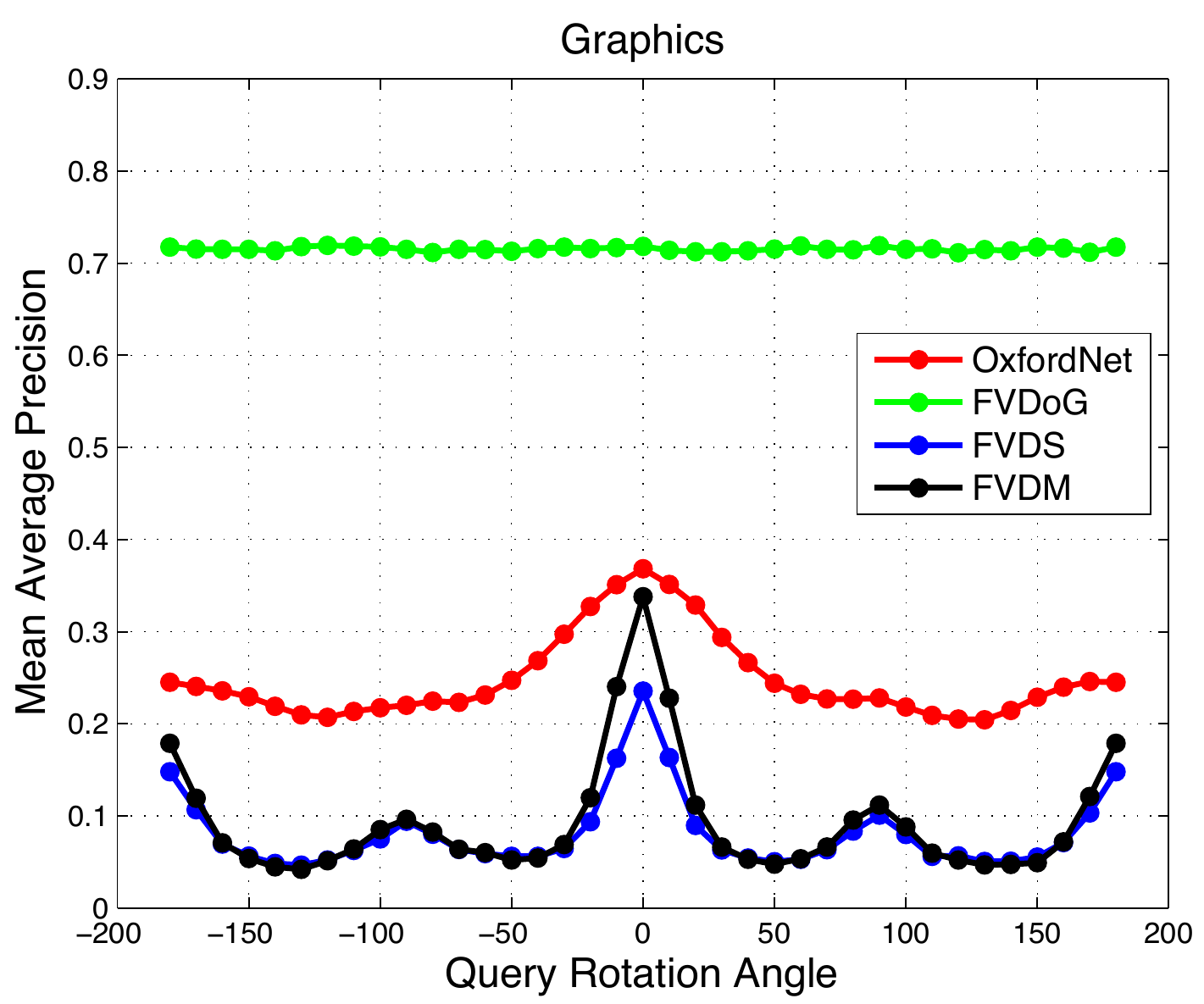}
	\end{tabular}
	\caption{\footnotesize Comparison of CNN and FV descriptors retrieval performance with rotated queries on {\it Holidays} and {\it Graphics} datasets (see experimental section for details on datasets).  FVDoG, FVDS, and FVDM are Fisher Vectors based on DoG interest points, Dense interest points at Single scale, and Dense interest points at Multiple scales respectively: all use the SIFT descriptor. FVDoG is robust to rotation, while CNN, FVDS and FVDM suffer a sharp drop in performance as query image is rotated.
		}
\label{fig:cnn_vs_fisher_rotation}
}
\vspace*{-3mm}	
\end{figure}

Unlike FVs based on interest point detectors like the Difference-of-Gaussian (DoG) detector, 
CNN does not have a built-in mechanism to ensure resilience to geometric transformations like scale and rotation.
In particular, the performance of CNN descriptors quickly degrade when the objects in the query and the database image are rotated or scaled differently.
To illustrate this, in Figure~\ref{fig:cnn_vs_fisher_rotation}, we show retrieval results when query images are rotated with respect to database images for descriptors: 
(a) Fisher Vectors based on Difference of Gaussian interest points and SIFT descriptors (FVDoG), 
(b) Fisher Vectors based on dense interest points and SIFT descriptors, at just one scale (FVDS), 
(c) Fisher Vectors based on dense interest points and SIFT descriptors, at multiple scales in the image (FVDM), 
and (d) CNN descriptors based on the first fully connected layer of {\it OxfordNet}~\cite{Simonyan2014}.
Schemes apart from FVDoG suffer a sharp drop in performance as geometric transforms are applied.
In this work, we focus on how to systematically incorporate groups of invariance into the CNN pipeline.

{\textbf{Hashing}.} 
In this work, we are focused on image instance retrieval with compact descriptors produced by unsupervised hashing on global descriptors.
Semantic image retrieval with supervised hashing is outside the scope of this work.
Examples of popular unsupervised hashing methods include Locality Sensitive Hashing (LSH)~\cite{LSH}, Iterative Quantization (ITQ)~\cite{ITQ}, 
Spectral Hashing (SH)~\cite{SpectralHashing} and Restricted Boltzmann Machines (RBM)~\cite{HintonScience,hintonSparsity}.
Gong et al. propose the popular ITQ~\cite{ITQ}.
ITQ first performs Principal Component Analysis (PCA) to reduce dimensionality, 
then applies rotations to distribute variance across dimensions, and finally binarizes each dimension according to its sign.
Besides hashing,
quantization based methods such as Product Quantization (PQ)~\cite{PQFisher,icml2014c2_zhangd14,Zhang_2015_CVPR} 
divide the raw descriptor into smaller blocks and vector quantization is performed on each block.
While this results in highly compact descriptors composed of sub-quantizer indices,
the resulting representation is not binary and cannot be compared with Hamming distance.

\begin{figure}
\centering
\includegraphics[width=0.9\textwidth]{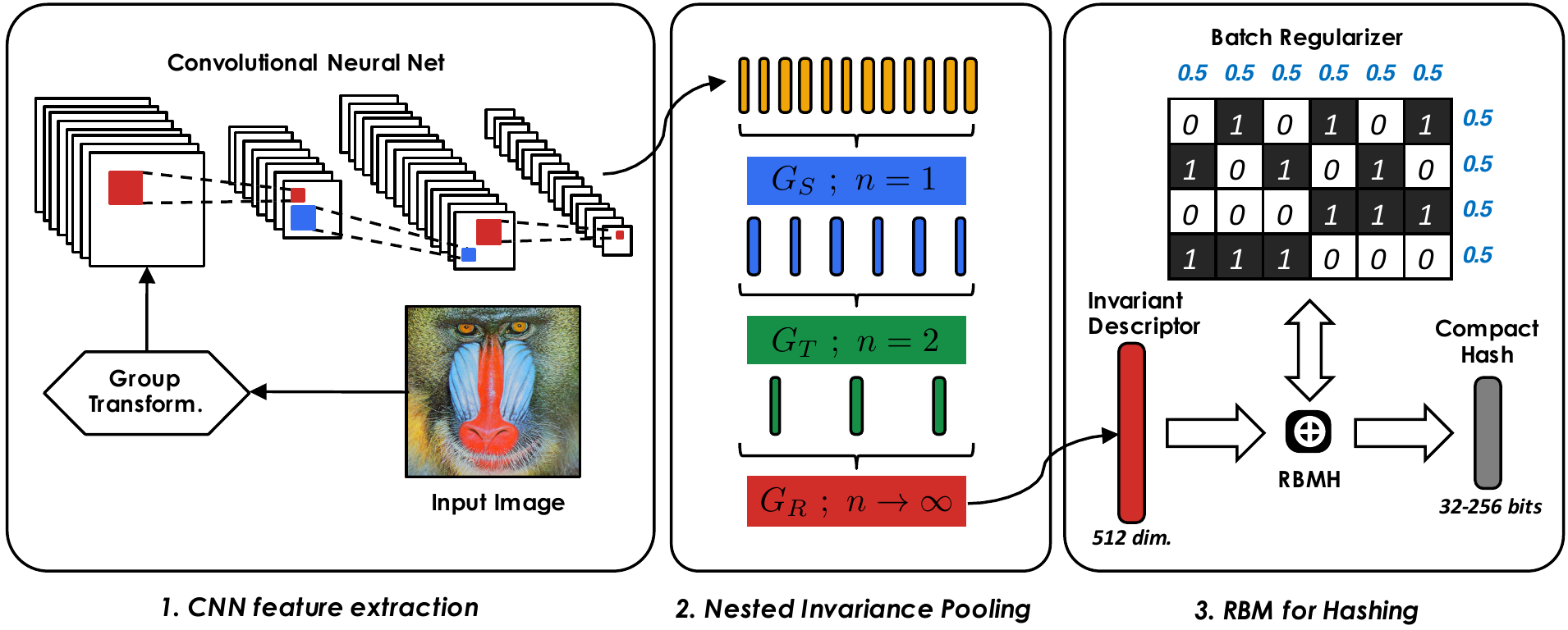}
\caption{\footnotesize 
Our proposed pipeline for image instance retrieval applies Nested Invariance Pooling (NIP) to produce robust and compact descriptors from CNNs followed by an RBM specially regularized for Hashing (RBMH).
}
\label{fig:method}
\vspace*{-3mm}
\end{figure}

\section{Contributions}

The goal of this work is the computation of very compact binary hashes for image instance retrieval.
To that end, we propose a multi-stage pipeline as shown on Figure~\ref{fig:method} with the following contributions:
\begin{itemize}
\item First, we propose Nested Invariance Pooling (NIP), a method to produce compact global image descriptors from visual representations extracted from CNNs.
Our method draws its inspiration from the \emph{i-theory}~\cite{itheory1,itheory2,itheory3}, a mathematical theory for computing group invariant transformations with feed-forward neural networks.
We specifically incorporate scale, translation and rotation invariance but the scheme can be extended to any arbitrary sets of transformations.
We also show that using moments of increasing order throughout nesting is important.
Resulting NIP descriptors are invariant to various types of image transformations and we show that the process significantly improves retrieval results while keeping dimensionality low (512 dimensions).
\item Then, the NIP descriptors are hashed to the target code size (32-256 bits) with a Restricted Boltzmann Machine (RBM).
We propose a novel batch-level regularization scheme specifically designed for the purpose of hashing, a scheme we refer to as RBMH from hereon. 
\item A thorough empirical evaluation with state-of-the-art shows that the results obtained both with the NIP descriptors and the NIP+RBMH hashes are consistently outstanding across a wide range of datasets.
To the best of our knowledge, the results reported at 128 bits hashes are the highest reported results in state-of-the-art literature.
\end{itemize}

\section{Method}
\label{sec:approach}

\begin{figure}
\centering
\includegraphics[width=0.9\textwidth]{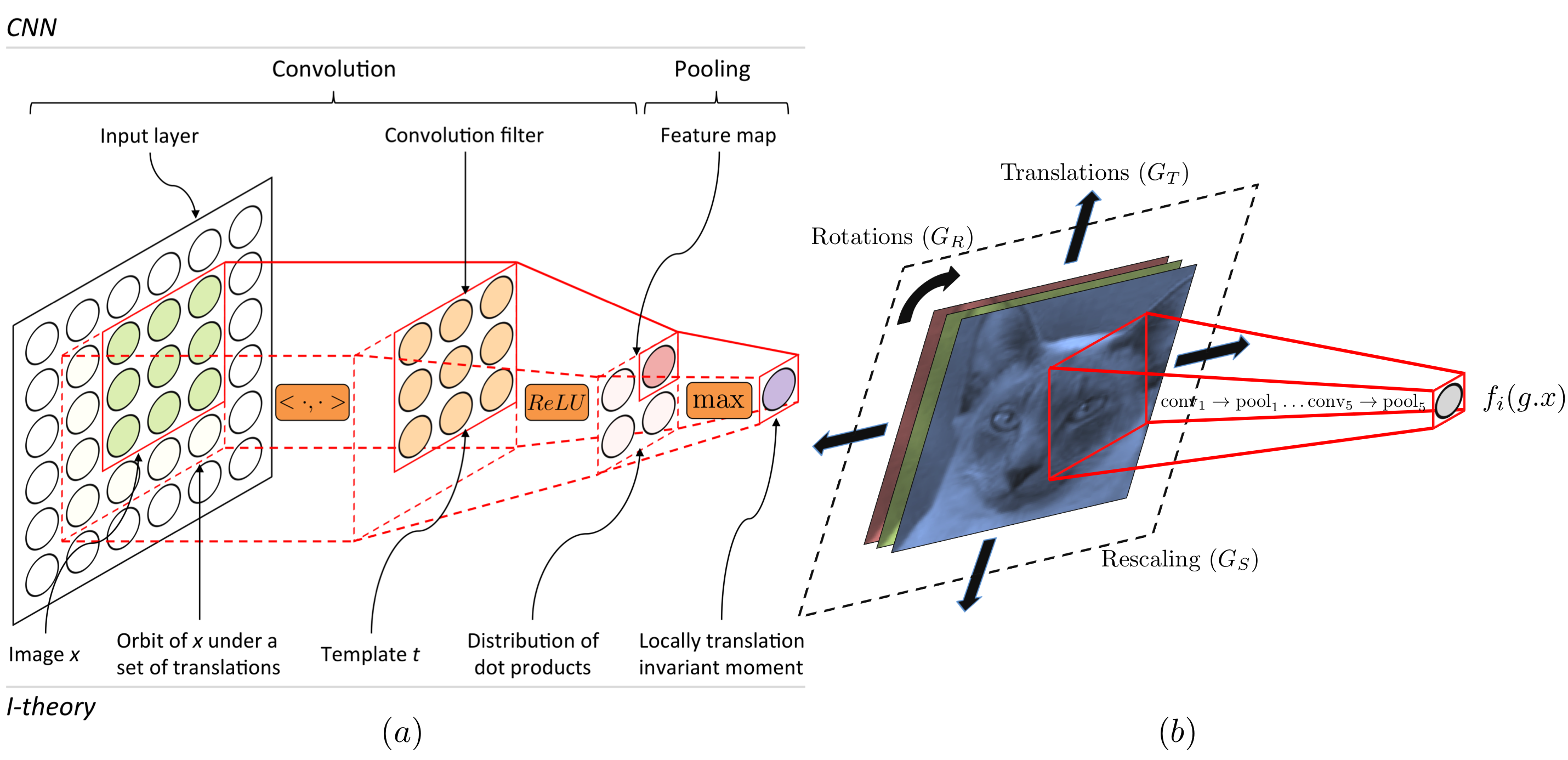}
\caption{\footnotesize (a) A single convolution-pooling operation from a CNN schematized for a single input layer and single output unit.
The parallel with \emph{i-theory} shows that the universal building block of CNNs is compatible with the incorporation of invariance to local translations of the input according to the theory.
(b) A specific succession of convolution and pooling operations learnt by the CNN (depicted in red) computes the \emph{pool5} feature $f_i$ for each feature map $i$ from the RGB image data.
A number of transformations $g$ can be applied to the input $x$ in order to vary the response $f_i(g.x)$.
}
\label{fig:framework}
\vspace*{-3mm}
\end{figure}

\subsection{Nested Invariance Pooling}
\label{sec:itheory}

Let an image $x \in E$ and a group $G$ of transformations acting over $E$ with group action $G \times E \rightarrow E$ denoted with a dot ($.$).
The orbit of $x$ by $G$ is the subset of $E$ defined as $O_x = \{ g.x \in E | g \in G \}$.
The orbit corresponds to the set of transformations of $x$ under groups such as rotations, translations and scale changes.
It can be easily shown that $O_x$ is globally invariant to the action of any element of $G$ and thus any descriptor computed directly from $O_x$ would be globally invariant to $G$.

The \emph{i-theory} builds invariant representations for a given object $x \in E$ in relation with a predefined template $t \in E$ from the distribution of the dot products $D_{x,t} = \{< g.x , t >  \in \mathbb{R} | g \in G \} = \{< x , g.t > \in \mathbb{R} | g \in G \}$ over the orbit. 
The following representation (for any $n \in \mathbb{N}^*$) is proven to have proper invariance and selectivity properties provided that the group is compact or locally compact:
\begin{align} 
&\mu_{G,t,n}(x) = \displaystyle \frac{1} {\int_G dg}\left(\int_G |<g.x, t>|^n dg \right)^{\frac{1}{n}} \label{eqn0}
\end{align}
Note that the sequence $(\mu_{G,t,n}(x))_{n \in \mathbb{N}^*}$ is analogous to a histogram.
In practice, the theory extends well (with approximate invariance) to non-locally compact groups and even to continuous non-group transformations (e.g. out-of-plane rotations, elastic deformations) provided that proper class-specific templates can be chosen~\cite{itheory1}.
Recent work on face verification \cite{poggioface} and music classification \cite{poggiomusic} apply the theory to non-compact groups with good results.

Popular CNN architectures for classification such as \emph{AlexNet}~\cite{AlexNet} and \emph{OxfordNet}~\cite{Simonyan2014} share a common building block: a succession of convolution-pooling operations designed to model increasingly high-level visual representations of the data.
As shown in Figure~\ref{fig:framework}~(a), the succession of convolution and pooling operations in a typical CNN is in fact a way to incorporate local translation invariance strictly compliant with the framework proposed by the \emph{i-theory}.
The network architecture provides the robustness as predicted by the \emph{i-theory}, while parameter tuning via back propagation ensures a proper choice of templates.

We build our NIP descriptors starting from the already locally robust \emph{pool5} feature maps of {\it OxfordNet}.
Global invariance to several transformation groups are then sequentially incorporated following the \emph{i-theory} framework.
The specific transformation groups considered in this study are translations $G_T$, rotations $G_R$ and scale changes $G_S$.
For every feature map $i$ of the \emph{pool5} layer ($0 \leq i < 512$), we denote $f_i(x)$ the corresponding unit's output.
As shown on Figure~\ref{fig:framework}~(b), transformations $g$ are applied on the input image $x$ varying the output of the \emph{pool5} feature $f_i(g.x)$.
Note that the transformation $f_i$ is non-linear due to multiple convolution-pooling operations thus not strictly a mathematical dot product but can still be viewed as an inner product.
Accordingly, the pooling scheme used by NIP with $G \in \{ G_T, G_R, G_S \}$ is:
\begin{align}
{\mathcal{X}}_{G,i,n}(x) = \displaystyle \frac{1} {\int_G dg}\left(\int_G f_i(g.x)^n dg \right)^{\frac{1}{n}} = \frac{1} {m}\left(\sum_{j=0}^{m-1} f_i(g_j.x)^n \right)^{\frac{1}{n}} \label{eqn0a}
\end{align}
when $O_x$ is discretized into $m$ samples.
The corresponding global image descriptors are obtained after each pooling step by concatenating the moments for the individual features:
\begin{align}
&{\mathcal{X}}_{G,n}(x) = ( {\mathcal{X}}_{G,i,n}(x) )_{0 \leq i < 512}
\end{align}

As shown in Equation~\ref{eqn0a}, the pooling operation has an order parameter $n$ defining the ``hardness'' of the pooling.
$n=1$ is average pooling while $n \rightarrow +\infty$ on the other extreme is max-pooling.
$n=2$ is analogous to standard deviation.
Subsequently, we refer to the moments for $n=1,2,+\infty$ as ${\mathcal{A}}_{G}$, ${\mathcal{S}}_{G}$ and ${\mathcal{M}}_{G}$.

Work on \emph{i-theory}~\cite{poggiomusic} has shown that it is possible to chain multiple types of group invariances one after the other~\cite{poggiomusic}.
We apply this principle on our NIP descriptors by making them invariant to several transformations.
For instance, following scale invariance with average ($n=1$) by translation invariance with hard max-pooling ($n \rightarrow +\infty$) is done by:
\begin{align} 
\displaystyle \max_{{g_t} \in G_T} \left( \frac{1}{\int_{g_s \in G_S} dg_s}\int_{g_s \in G_S} f_i(g_tg_s.x) dg_s \right) =  \max_{j \in [0,m_t - 1]} \left( \frac{1}{m_s}\sum_{i=0}^{m_s - 1} f_i(g_{t,j}g_{s,i}t.x)\right) 
\end{align}
Operations are sometimes commutable (e.g. $\mathcal{A}_G$ and $\mathcal{A}_{G'}$) and sometimes not (e.g. $\mathcal{A}_G$ and $\mathcal{M}_{G'}$) depending on the specific combination of moments so the sequence of transformations does matter for NIP.
The hardness parameter $n$ must also be chosen carefully.
Empirically, we found pooling progressively with increasing moments (e.g. ${\mathcal{A}}_{G}$, then ${\mathcal{S}}_{G}$, then ${\mathcal{M}}_{G}$) to work well as presented in the experiments section.

\begin{figure}[ht]
\centering
\includegraphics[width=.5\textwidth]{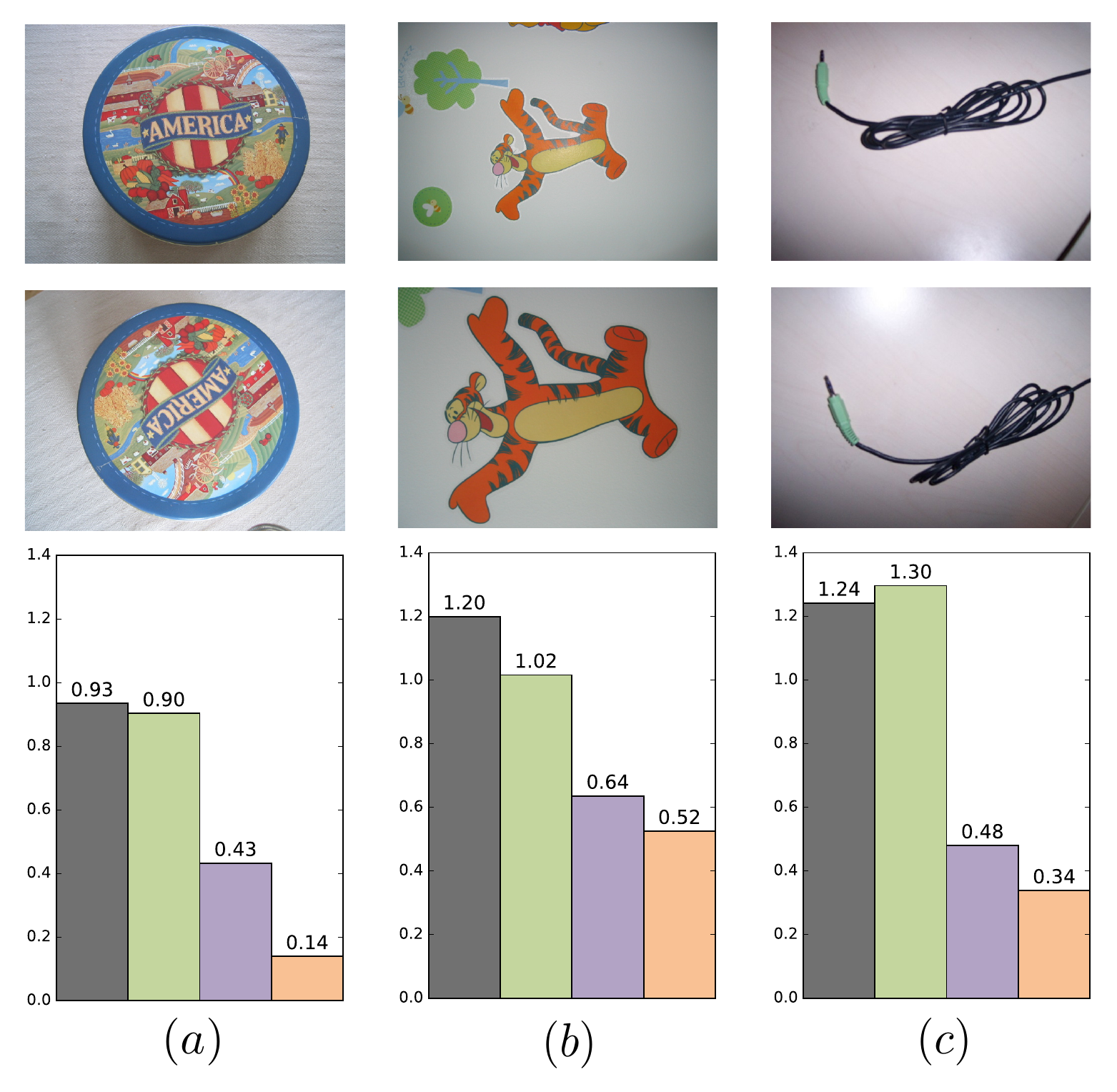}
\begin{tabular}{@{}c@{}cc@{}cc@{}cc@{} c}
\includegraphics[width=.6cm]{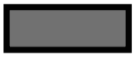}&\emph{pool5}&
\includegraphics[width=.6cm]{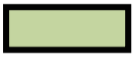}&$\mathcal{A}_{G_S}$&
\includegraphics[width=.6cm]{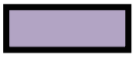}&$\mathcal{A}_{G_S}$-$\mathcal{A}_{G_T}$&
\includegraphics[width=.6cm]{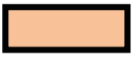} &$\mathcal{A}_{G_S}$-$\mathcal{A}_{G_T}$-$\mathcal{A}_{G_R}$
\end{tabular}
\caption{\footnotesize Distances for 3 matching pairs from the \emph{UKB} dataset. 
For each pair, 4 pairwise distances ($L_2$-normalized) are computed corresponding to the following descriptors: 
\emph{pool5}, $\mathcal{A}_{G_S}$, $\mathcal{A}_{G_S}$-$\mathcal{A}_{G_T}$ and  $\mathcal{A}_{G_S}$-$\mathcal{A}_{G_T}$-$\mathcal{A}_{G_R}$. 
Adding scale invariance makes the most difference on (b), translation invariance on (c), and rotation on (a) which is consistent with the scenarios suggested by respective image pairs.}
\label{fig:dists}
\vspace*{-3mm}
\end{figure}

Figure~\ref{fig:dists} provides an insight on how adding different types of invariance with our NIP scheme will affect the matching distance on different image pairs of matching objects.
With the incorporation of each new transformation group, we notice that the relative reduction in matching distance is the most significant with the image pair which is the most affected by the transformation group.

\subsection{Restricted Boltzmann Machine for Hashing}
\label{sec:rrbm}

The NIP descriptors are subsequently hashed to the target dimensionality with an RBM layer. 
The motivation is to obtain mutually independent dimensions while distributing variance evenly across them in a way similar to ITQ~\cite{ITQ}.
The main originality of our RBM is its batch-level regularization scheme which is specifically designed for hashing. 
We subsequently refer to this variant as RBMH.

An RBM is a bipartite Markov random field with the input layer $x\in{R}^{I}$ connected to a latent layer $z\in{R}^{J}$ via a set of undirected weights $W\in{R}^{I J}$. 
The input and latent layers are also parameterised by their corresponding biases $c$ and $b$, respectively.
Since the units within a layer are conditionally independent pairwise, the activation probabilities of one layer can be sampled by fixing the states of the other layer, and using distributions given by logistic functions (a sigmoid activation function is chosen since binary hashes are desired):
\begin{equation}
{P}(z_j| x) = 1/(1+\exp(-{w}_{j} x - b_j)),
\label{eq:alt_1}
\end{equation}
\begin{equation}
{P}(x_i| z) = 1/(1+\exp(-{w}_{i}^{{\top}} {z} - c_i)).
\label{eq:alt_2}
\end{equation}
As a result, alternating Gibbs sampling can be performed between the two layers. 
The sampled states are used to update the parameters $\{{W},{b},{c}\}$ through batch gradient descent 
using the contrastive divergence algorithm~\cite{hintonCD} to approximate the maximum likelihood of the input distribution.
The hashed descriptors are obtained by binarizing the latent units at 0.5.

Proper regularization is key during the training of RBM.
The popular RBM proposed by Nair and Hinton~\cite{hintonSparsity} encourages latent representations to be sparse.
This improves separability which is desirable for classification task.
For hashing, it is desirable to encourage the representation to make efficient use of the limited latent subspace.
RBMH achieves this goal by controlling sparsity in a way to maximize the entropy not only within every hash but also between the same bit of different hashes.
This effectively encourages (a) half the bits to be active for a given hash, and (b) each hash bit to be equiprobable across images.
We introduce a regularization term at the batch as in~\cite{hanlinSparsity}.
For a batch $B$, we define a regularization term:
\begin{align}
h(B) = \sum_{x_{\alpha}\in B} \sum_{j\alpha} t_{j\alpha} \log z_{j\alpha} +(1-t_{j\alpha}) \log(1-z_{j\alpha}),
\end{align}
where $t_{\alpha}$ are the target activations $Z_{\alpha}$ for each data sample $X_{\alpha}$. 
We choose the $t_{j\alpha}$ such that each $\{t_{j\alpha}\}_{j}$ for fixed $\alpha$ and each $\{t_{j\alpha}\}_{\alpha}$ for fixed $j$ is distributed according to the uniform distribution ${U}(0,1)$ effectively maximizing entropy.
The overall objective function becomes:
\begin{align}
\underset{\left\{W,b,c\right\}}
{\arg\min} - \sum_{\alpha}\log \bigg(\sum_{z_{\alpha}\in{B}}{P}({x}_{\alpha},{z}_{\alpha})+\lambda h(B)\bigg),
\label{eq:sparseproblem}
\end{align}
with $\lambda$ the regularization constant.

\begin{figure}[ht]
\centering
\begin{tabular}{@{}c@{} @{}c@{}}
\includegraphics[width=.4\textwidth]{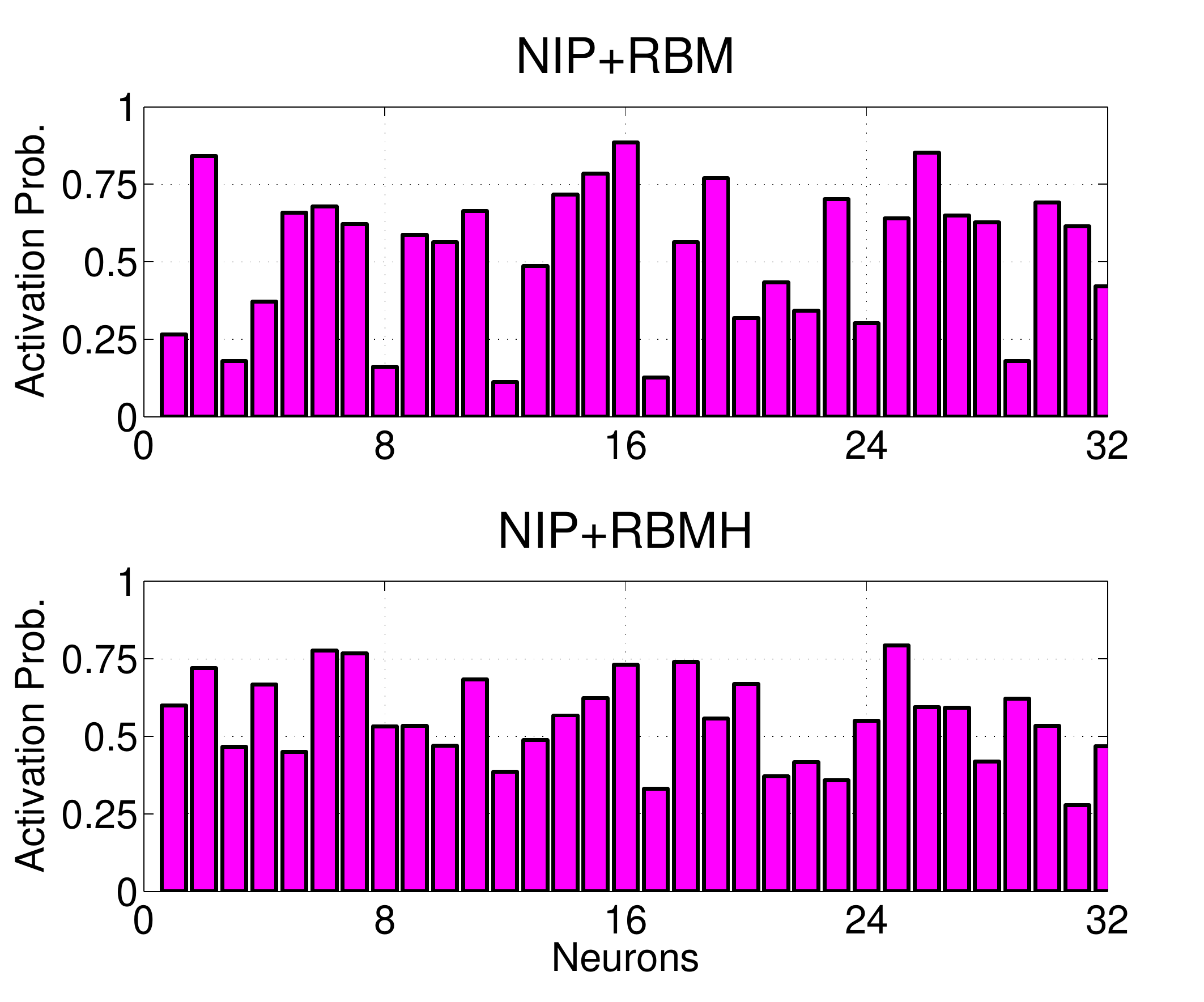} &
\includegraphics[width=.4\textwidth]{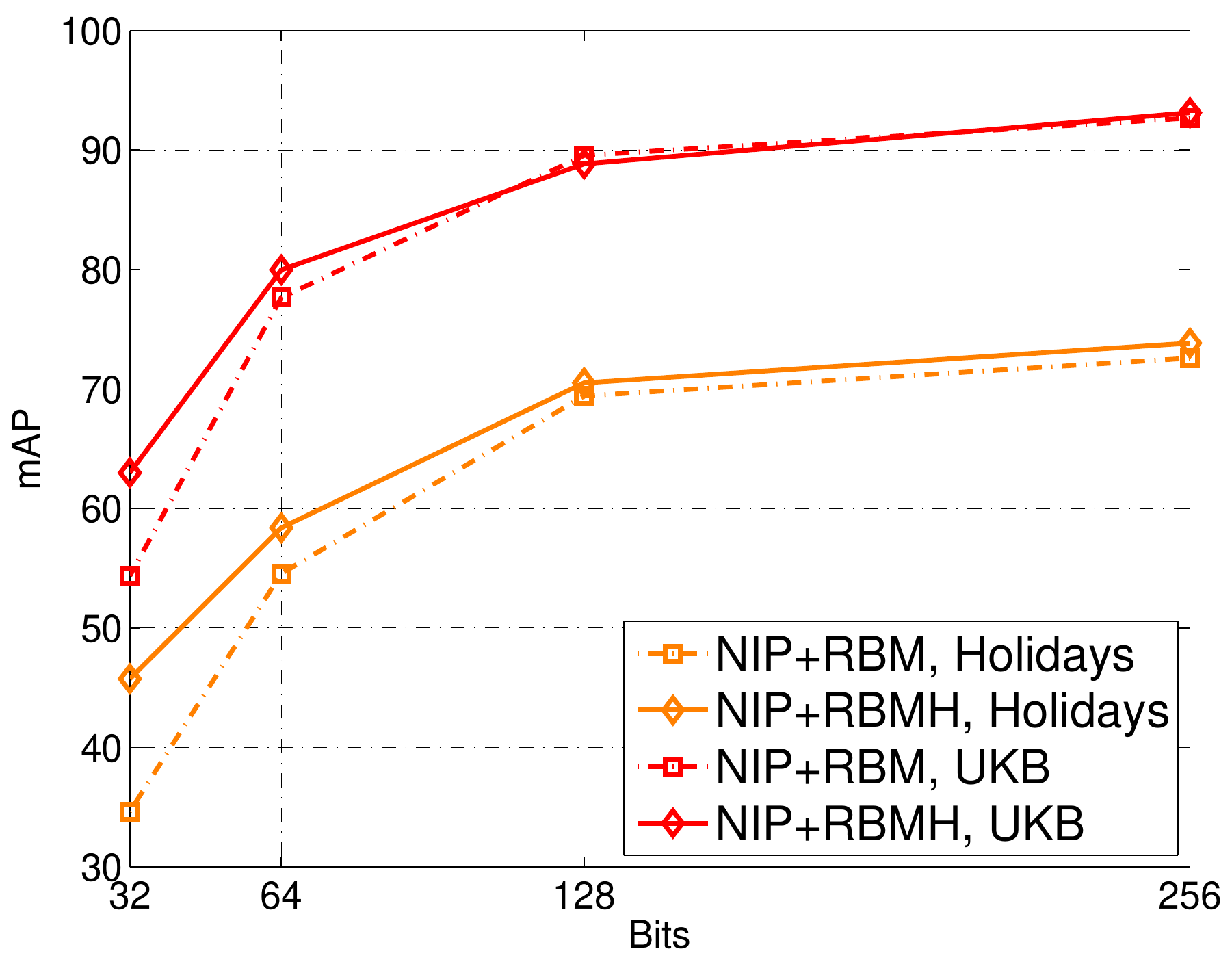} \\
	(a) & (b) \\
\end{tabular}
\caption{\footnotesize 
(a) Activation probabilities of hash bits between RBMH and the RBM proposed by Nair\&Hinton~\cite{hintonSparsity}.
We compute the statistics with 32 bits binary hashes on {\it Holidays} dataset (1491 images in total).
(b) Comparison of our RBMH with RBM~\cite{hintonSparsity} in terms of mAP on {\it Holidays} and {\it UKB}.
Both schemes are built upon the best NIP descriptors.
RBMH outperforms RBM at very low code sizes.
}
\label{fig:sparsity}
\vspace*{-3mm}
\end{figure}

Figure~\ref{fig:sparsity} shows the activation probabilities of the hash bits between RBMH and the RBM proposed by Nair and Hinton~\cite{hintonSparsity}.
The comparison is for 32-bits hashes.
In Figure~\ref{fig:sparsity} (a), the mean probability of activation is nearly 0.5 in both cases.
Nevertheless, we can see that probabilities are much more evenly distributed across bits with RBMH.
In Figure~\ref{fig:sparsity} (b), retrieval results on {\it UKB} and {\it Holidays} show that the RBMH is able to 
outperform the standard RBM specially at lower code sizes (32 or 64 bits).

\section{Experiments}
\label{sec:exps}

\subsection{Evaluation Framework}
\label{sec:dataset}

We evaluate the performances on 4 popular datasets for image instance retrieval: 
(1) \textbf{Holidays}.
The INRIA Holidays dataset~\cite{Jegou08} consists of outdoor holiday pictures. 
There are 500 queries and 991 database images.
(2) \textbf{UKB}.
The University of Kentucky Benchmark dataset~\cite{Nister06} consists of 2550 groups of common objects, 4 images per object.
All 10200 images are used as queries.
(3) \textbf{Oxford5K}.
The Oxford buildings dataset~\cite{Philbin07} consists of 5063 images representing landmark buildings in Oxford. 
The query set contains 11 different landmarks, each represented by 5 queries.
(4) \textbf{Graphics}.
The Graphics dataset is part of the Stanford Mobile Visual Search dataset~\cite{SVMSDataSet}, 
which was used in the MPEG standard titled Compact Descriptors for Visual Search (CDVS)~\cite{MPEGDataset2}.
This dataset contains objects like CDs, DVDs, books, prints, business cards.
There are 500 unique objects, 1500 queries, and 1000 database images.
Note that {\it Graphics} is different from the other datasets as it contains images of rigid objects captured under widely varying scale and rotation changes. 

For large-scale experiments,
we present results on the 4 datasets combined with the 1 million MIR-FLICKR distractor images~\cite{mirflickr}.
Most schemes, including our approach, require an unsupervised training step.
We train on a randomly sampled set of 150K images from the 1.2 million {\it ImageNet} dataset~\cite{DengImagenet}.
No class labels are used in this work.

For the starting global descriptor representation, we use the \emph{pool5} layer from the 16-layer \emph{OxfordNet} \cite{Simonyan2014}, 
which is widely adopted in instance retrieval literature~\cite{sharif2015baseline,babenko2,ToliasSJ15}.
The input image size for {\it OxfordNet} is fixed at $224 \times 224$.
The dimensionality of the \emph{pool5} descriptor is $25088$, organized as $512$ feature maps of size $7 \times 7$.

For rotation invariance, rotated input images are padded with the mean pixel value computed from the ImageNet dataset.
The step size for rotations is 10 degrees yielding 36 rotated images per orbit.
For scale changes, 10 different center crops are considered varying from 50\% to 100\% of the total image.
For translations, the entire feature map is used for every feature, resulting in an orbit size of $7 \times 7 = 49$.

For retrieval with floating point descriptors, $L_2$ normalization is first applied followed by L2 distance computation.
For retrieval with binary descriptors, we use hamming distance computation.
We evaluate retrieval results with mean Average Precision (mAP).
To be consistent with the literature, 4$\times$ Recall @ $R=4$ is provided for {\it UKB}.
\begin{table}[ht]
\caption{\footnotesize Retrieval results (mAP) for different sequences of transformation groups and moments.
For {\it UKB}, 4$\times$ Recall @ $R=4$ is shown between parentheses.
$G_T$, $G_R$, $G_S$ denote the groups of translations, rotations and scale changes respectively.
Note that averages commute with other averages so the sequence order of the composition does not matter when only averages are involved.
Best results are achieved by choosing specific moments.
$\mathcal{A}_{G_S}$-$\mathcal{S}_{G_T}$-$\mathcal{M}_{G_R}$ corresponds to the best average performer.
\emph{fc6} and \emph{pool5} are provided as a baseline.}
\label{tab:res} 
\centering
\ra{1.2}
{\footnotesize \singlespacing 
\begin{tabular}{@{}lrcccc@{}}
\toprule
\multirow{2}{*}{\sc Sequence} & \multirow{2}{*}{\sc Dims} & \multicolumn{4}{c}{\sc Dataset} \\
\cmidrule{3-6}
 & & Oxford5K & Holidays & UKB & Graphics\\
\midrule
\emph{pool5} & 25088 & 0.427 & 0.707 & 0.823(3.11) & 0.315\\
\emph{fc6} & 4096 & 0.461 & 0.782 & 0.910(3.50) & 0.312\\
\midrule
$\mathcal{A}_{G_T}$ & 512 & 0.477 & 0.800 & 0.924(3.56) & 0.322\\
$\mathcal{A}_{G_R}$ & 25088 & 0.462 & 0.779 & 0.954(3.72) & 0.500\\
$\mathcal{A}_{G_S}$ & 25088 & 0.430 & 0.716 & 0.828(3.12) & 0.394\\
$\mathcal{A}_{G_T}$-$\mathcal{A}_{G_R}$ & 512 & 0.418 & 0.796 & 0.955(3.73) & 0.417\\
$\mathcal{A}_{G_T}$-$\mathcal{A}_{G_S}$ & 512 & 0.537 & 0.811 & 0.931(3.61) & 0.430\\
$\mathcal{A}_{G_R}$-$\mathcal{A}_{G_S}$ & 25088 & 0.494 & 0.815 & 0.959(3.75) & 0.552\\
$\mathcal{A}_{G_T}$-$\mathcal{A}_{G_R}$-$\mathcal{A}_{G_S}$ & 512 & 0.484 & 0.833 & 0.971(3.82) & 0.509\\
\midrule
$\mathcal{A}_{G_S}$-$\mathcal{S}_{G_T}$-$\mathcal{M}_{G_R}$ & 512 & \textbf{0.592} & \textbf{0.838} & \textbf{0.975(3.84)} & \textbf{0.589} \\
\bottomrule
\end{tabular}
}
\vspace*{-3mm}
\end{table}

\subsection{Results}
\label{sec:results}

\textbf{Evaluation of NIP descriptors}.
As shown in Table~\ref{tab:res}, we first study the effects of incorporating various transformation groups and using different moments on NIP descriptors.
We present results for all possible combinations of transformation groups for average pooling 
(order does not matter as averages commute) and for the single best performer which is $\mathcal{A}_{G_S}$-$\mathcal{S}_{G_T}$-$\mathcal{M}_{G_R}$ (order matters).
First, we point out the effectiveness of the \emph{pool5} layer.
Although it performs notably worse than \emph{fc6} as-is, 
a simple average pooling over the space of translations \emph{$\mathcal{A}_{G_T}$} makes it both better and 8 times more compact than \emph{fc6}.
Similar observations have also been reported by~\cite{babenko2,Razavian15}.
Second, on average, accuracy significantly increases with the number of transformation groups involved.
Third, choosing statistical moments different than averages further improve the retrieval results.
In Table~\ref{tab:res}, we observe that $\mathcal{A}_{G_S}$-$\mathcal{S}_{G_T}$-$\mathcal{M}_{G_R}$ performs 
significantly better than $\mathcal{A}_{G_T}$-$\mathcal{A}_{G_R}$-$\mathcal{A}_{G_S}$.
Notably, the best combination corresponds to an increase in the orders of the moments: 
$\mathcal{A}$ being a first-order moment, $\mathcal{S}$ second order and $\mathcal{M}$ of infinite order.
A different way of stating this is that a more invariant representation requires higher and higher orders of pooling.

\begin{table}[ht]
\caption{\footnotesize Retrieval performance comparing NIP to other state-of-the-art methods.
We include results in recent papers with comparable dimensionality of descriptors reported in those papers.
L2 distance is used for all methods.}
\label{tab:stateoftheart} 
\centering
\ra{1.2}
{\footnotesize \singlespacing 
\begin{tabular}{@{}lrccc@{}}
\toprule
\multirow{2}{*}{\sc Method} & \multirow{2}{*}{\sc Dims} & \multicolumn{3}{c}{\sc Dataset} \\
\cmidrule{3-5}
&  & Oxford5K & Holidays & UKB\\
\midrule
{T-embedding~\cite{TEDA}} & 1024 & 0.560 & 0.720 & 3.51\\
{T-embedding~\cite{TEDA}} & 512 & 0.528 & 0.700 & 3.49\\
{FV+Proj~\cite{FisherColor}} & 512 & - & 0.789 & 3.36\\
\midrule
{{FC+PCAWhitening}~\cite{Razavian2014}} & 500 & 0.322 & 0.642 & -\\
{FC+VLAD+PCA~\cite{NewPaperA}} & 512 & - & 0.784 & -\\
{FC+Finetune+PCAWhitening~\cite{Yandex}} & 512 & 0.557 & 0.789 & 3.30\\
{Conv+MaxPooling~\cite{sharif2015baseline}} & 256 & 0.533 & 0.716 & -\\
{FV+FC+PCAWhitening~\cite{Perronnin_2015_CVPR}} & 512 & - & 0.827 & 3.37\\
{Conv+SPoC+PCAWhitening~\cite{babenko2}} & 256 & 0.589 & 0.802 & 3.65\\
{R-MAC+PCAWhitening~\cite{ToliasSJ15}} & 512 & \textbf{0.668} & - & -\\
{R-MAC+PCAWhitening~\cite{ToliasSJ15}} & 256 & 0.561 & - & -\\
\midrule
{NIP (Ours)} & 512 & 0.592 & \textbf{0.838} & \textbf{3.84}\\
{NIP+PCAWhitening (Ours)} & 256 & \textbf{0.609} & \textbf{0.836} & \textbf{3.83}\\
\bottomrule
\end{tabular}
}
\vspace*{-3mm}
\end{table}

Overall, $\mathcal{A}_{G_S}$-$\mathcal{S}_{G_T}$-$\mathcal{M}_{G_R}$ improves results 
over starting representation \emph{pool5} by 39\% ({\it Oxford5K}) to 87\% ({\it Graphics}) depending on the dataset.
Better improvements with {\it Graphics} can be explained with the presence of many rotations in the dataset 
(smaller objects taken under different angles) while {\it Oxford5K} consisting mainly of upright buildings is less significantly helped by incorporating rotation invariance.

\textbf{Comparing NIP with state-of-the-art} including variants of VLAD/FV~\cite{TEDA,FisherColor}, 
deep descriptors~\cite{Yandex,sharif2015baseline,babenko2,ToliasSJ15} and 
descriptors combining deep CNN and VLAD\slash{}FV~\cite{NewPaperA,Perronnin_2015_CVPR}.
As shown in Table~\ref{tab:stateoftheart}, we observe that 512-D NIP descriptors largely outperform most state-of-the-art methods with 512 or higher dimensions, on all datasets.
Following~\cite{sharif2015baseline,babenko2,ToliasSJ15}, we also perform PCA whitening to reduce the dimensionality of NIP to 256.
One can see that the 256-D NIP descriptors yield superior performance to~\cite{sharif2015baseline,babenko2,ToliasSJ15} on all datasets.

First, we compare NIP to the most related papers~\cite{sharif2015baseline,babenko2,ToliasSJ15} 
which propose 256-D deep descriptors by aggregating convolutional features with various pooling operations~\footnote{Note that~\cite{sharif2015baseline,babenko2,ToliasSJ15} extract deep descriptors from images with size larger than $576 \times 576$, 
while we use $224 \times 224$ in this work.
As shown in~\cite{Yandex}, there is potential improvement if larger image size adopted in deep descriptors extraction.}. 
\cite{sharif2015baseline,babenko2,Razavian15} can be considered a special case of our work, 
with just one layer of pooling, which only provided limited levels of translation invariance.
The very recently proposed Regional Maximum Activation of Convolutions (R-MAC)~\cite{ToliasSJ15} 
reports outstanding results on building dataset {\it Oxford5K} with very small dimensionality (e.g. 0.668 mAP for 512-D R-MAC and 0.561 mAP for 256-D R-MAC).
The authors propose a fast R-CNN type pooling~\cite{girshick15fastrcnn}, 
which is effective when the object of interest is in a small portion of the image.
Such an approach will be less effective when the object of interest is affected by groups of distortions like rotation and perspective, and located at the centre of the image.  
Here, we observe that nested pooling over many types of distortions with progressively increasing moments 
is essential to achieving geometric invariance and high retrieval performance with low dimensional descriptors.
Besides, we argue that the technique proposed in~\cite{ToliasSJ15} can be incorporated with NIP to further improve performance.

\begin{figure}[ht]   
    \begin{minipage}[b]{.72\textwidth}
    \centering
      \begin{tabular}{cc}
        \includegraphics[width=.5\textwidth]{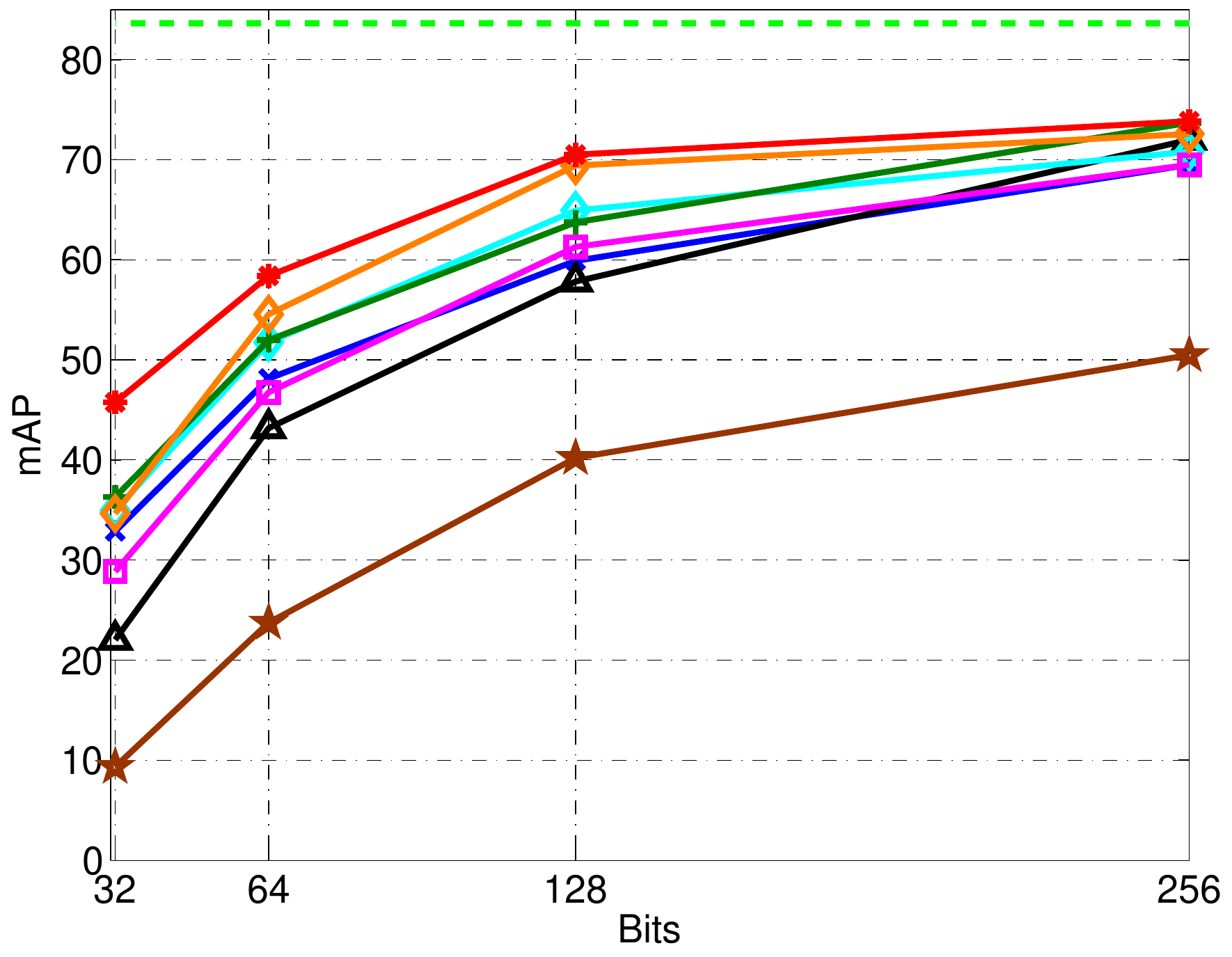} & 
        \includegraphics[width=.5\textwidth]{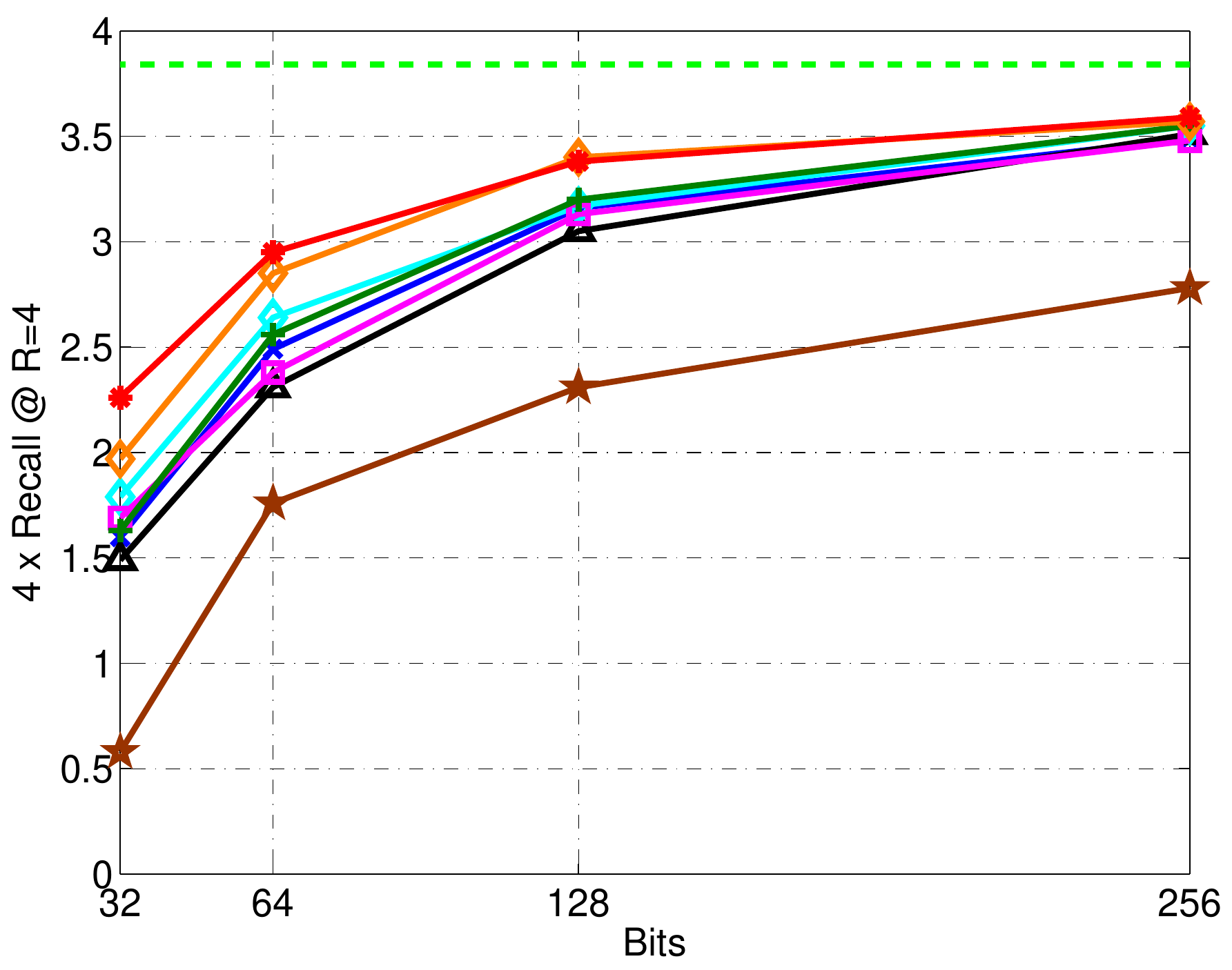} \\ 
        (a) Holidays & (b) UKB \\        
		\includegraphics[width=.5\textwidth]{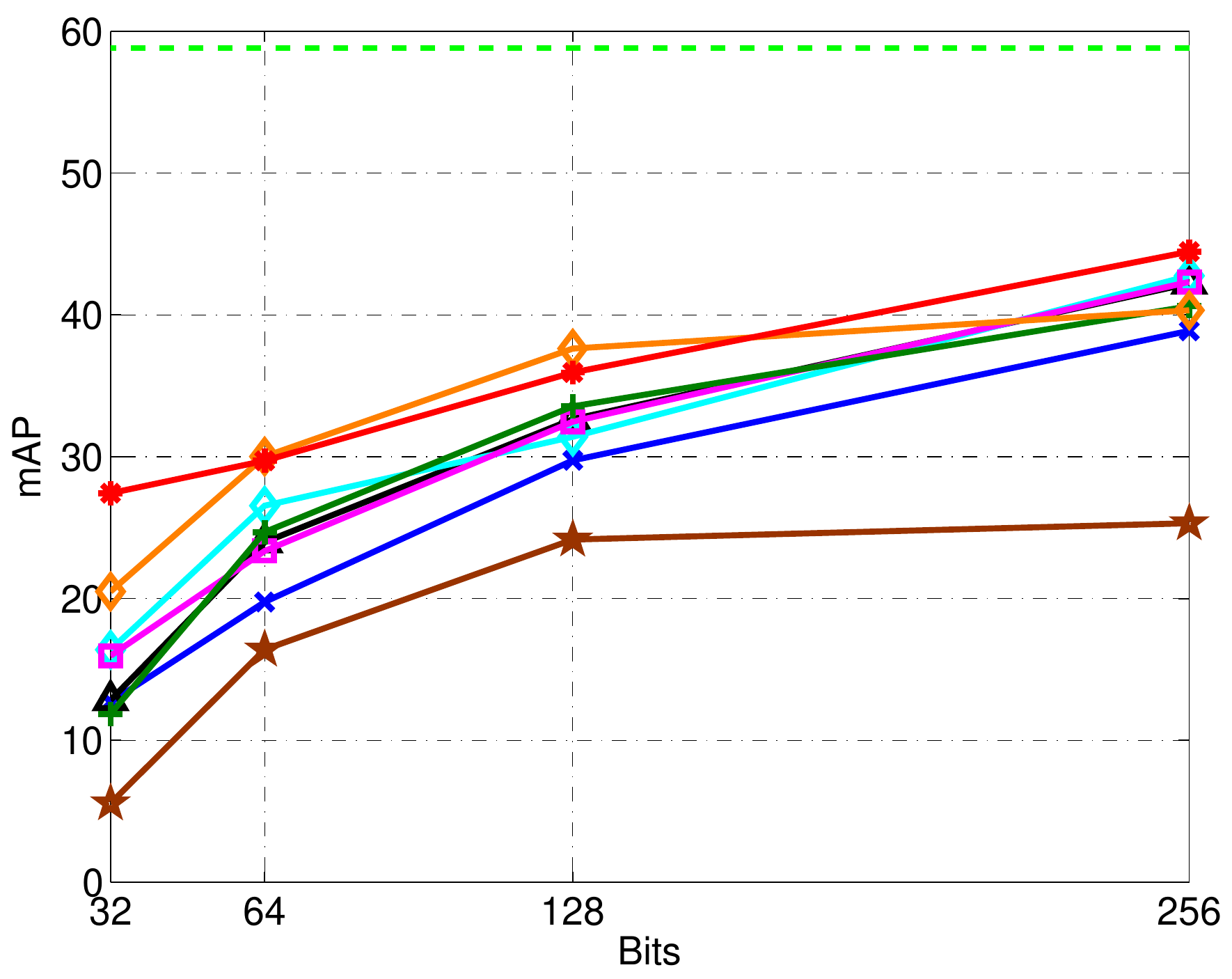} &
		\includegraphics[width=.5\textwidth]{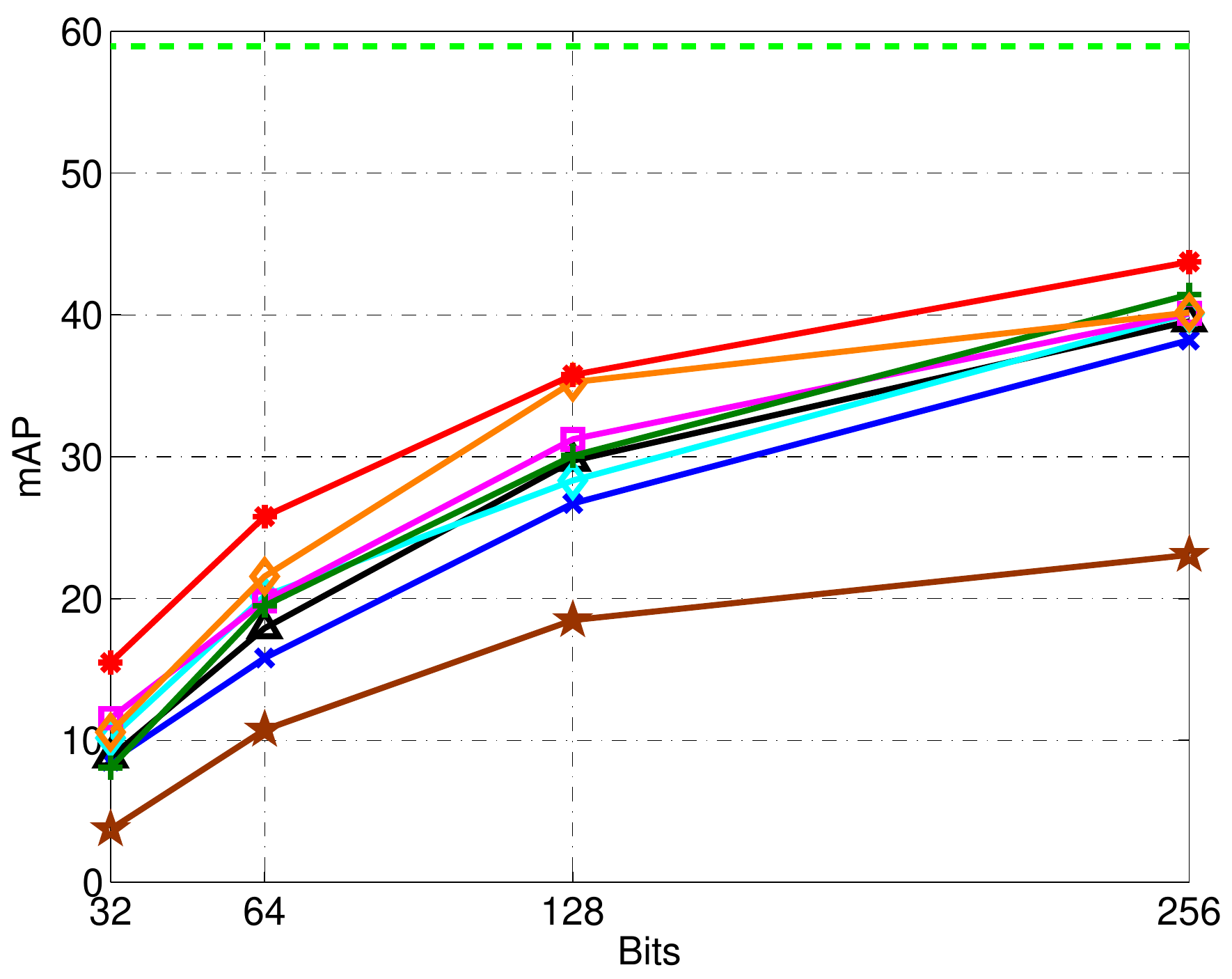} \\
		(c) Oxford5K & (d) Graphics        
      \end{tabular}
    \end{minipage}%
    \begin{minipage}[b]{.2\textwidth}
        \centering
        \includegraphics[width=0.85\textwidth]{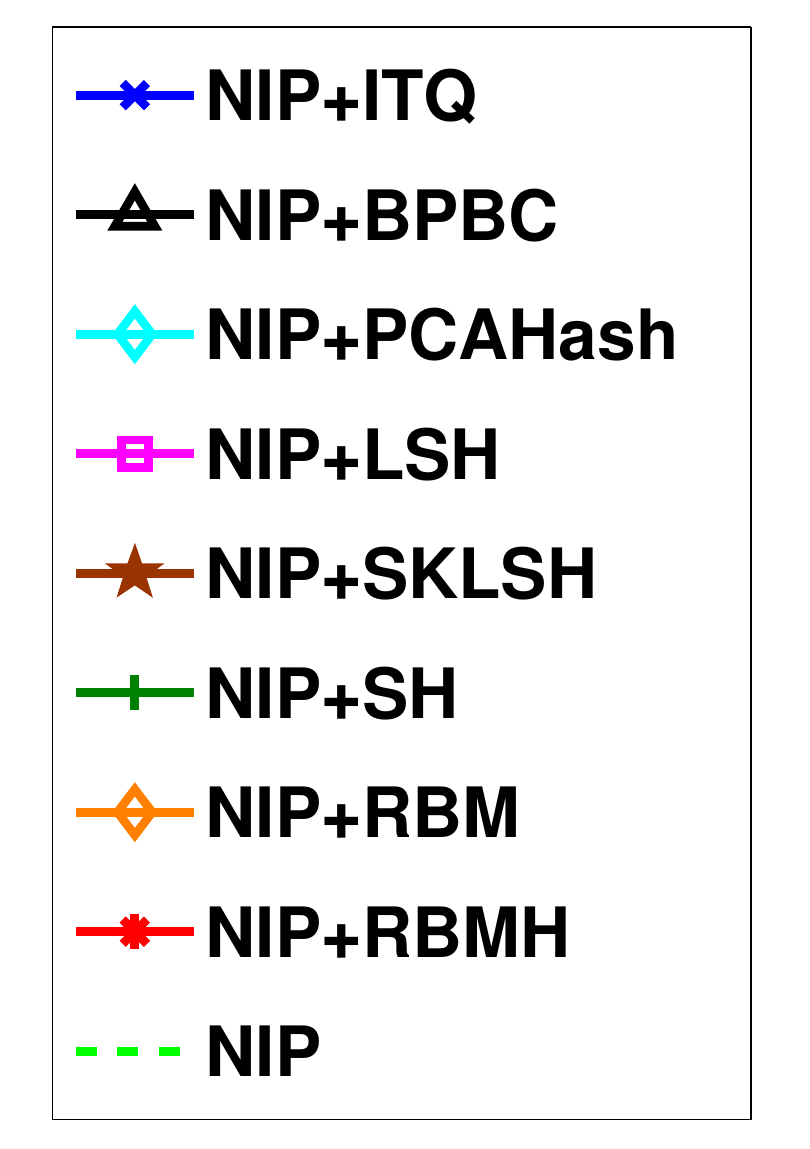}
    \end{minipage}%
\caption{\footnotesize Comparison of RBMH with other hashing methods on 4 benchmark datasets. All methods are built upon the best NIP descriptors.
To examine the effect of compression, we also present retrieval results using uncompressed NIP descriptors.}
\label{fig:reshashsmall}
\vspace*{-3mm}
\end{figure}

Next, we note that~\cite{sharif2015baseline} reports better results on {\it Holidays} (0.881 mAP) and {\it Oxford5K} (0.844 mAP), 
with very high-dimensional descriptors (from 10K to 100K).
These very high dimensional descriptors are obtained by combining CNN descriptors with spatial max pooling~\cite{Razavian15}.
In contrast, our results are generated using only 256 to 512 dimensional descriptors.

\begin{table}[ht]
\caption{\footnotesize Retrieval performance comparing NIP+RBMH to other state-of-the-art methods at small codesizes (from 32 to 512 bits).
ADC denotes asymmetric distance computation~\cite{PQFisher,asdembedding}.}
\label{tab:stateofthearthash} 
\centering
\ra{1.2}
{\footnotesize \singlespacing 
\begin{tabular}{@{}lrrccc@{}}
\toprule
\multirow{2}{*}{\sc Method} & \multirow{2}{*}{{\minitab[r]{\sc Dims \\ (size in bits)}}} & \multirow{2}{*}{\sc Dist.} & \multicolumn{3}{c}{\sc Dataset} \\
\cmidrule{4-6}
& & & Oxford5K & Holidays & UKB\\
\midrule
{Binarized FV~\cite{Perronnin_CVPR_10}} & 520(520) & Cosine & - & 0.460 & 2.79\\
{FV+SSH~\cite{asdembedding}} & 256(256) & ADC & - & 0.544 & 3.08\\
{FV+SSH~\cite{asdembedding}} & 128(128) & ADC & - & 0.499 & 2.91\\
{FV+SSH~\cite{asdembedding}} & 32(32) & ADC & - & 0.334 & 2.18\\
{FV+PQ~\cite{PQFisher}} & 128(128) & ADC & - & 0.506 & 3.10\\
{VLAD+PQ~\cite{icml2014c2_zhangd14}} & 128(128) & L2 & - & 0.586 & 2.88\\
{VLAD+CQ~\cite{icml2014c2_zhangd14}} & 128(128) & L2 & - & 0.644 & 3.19\\
{VLAD+SQ~\cite{Zhang_2015_CVPR}} & 128(128) & L2 & - & 0.639 & 3.06\\
\midrule
{FC+Finetune+PCAWhitening~\cite{Yandex}} & 16(512) & L2 & 0.418 & 0.609 & 2.41\\
{Conv+MaxPooling~\cite{sharif2015baseline}} & 256(256) & Cosine & 0.436 & 0.578 & -\\
\midrule
{Binarized NIP (Ours)} & 512(512) & Hamming & \textbf{0.477} & \textbf{0.781} & \textbf{3.70}\\
{NIP+RBMH (Ours)} & 256(256) & Hamming & \textbf{0.445} & \textbf{0.739} & \textbf{3.59}\\
{NIP+RBMH (Ours)} & 128(128) & Hamming & \textbf{0.359} & \textbf{0.705} & \textbf{3.38}\\
{NIP+RBMH (Ours)} & 32(32) & Hamming & \textbf{0.274} & \textbf{0.458} & \textbf{2.26}\\
\bottomrule
\end{tabular}
}
\vspace*{-3mm}
\end{table}

\textbf{Evaluation of NIP+RBMH binary hashes}.
NIP+RBMH binary hashes are produced by feeding invariant NIP descriptors into the proposed RBM hashing layer.
Small scale retrieval results with NIP+RBMH are shown in Figure~\ref{fig:reshashsmall}.
We compare NIP+RBMH to other popular unsupervised hashing methods at code sizes from 32 to 256 bits,
including {\it ITQ}~\cite{ITQ}, Bilinear Projection Binary Codes ({\it BPBC})~\cite{BPBC}, {\it PCAHash}~\cite{ITQ}, {\it LSH}~\cite{LSH}, {\it SKLSH}~\cite{SKLSH}, 
{\it SH}~\cite{SpectralHashing} and {\it RBM}~\cite{HintonScience,hintonSparsity}.
We used the software provided by the authors in~\cite{ITQ,HintonScience} to generate results for the baseline hashing methods.
In addition, we also include the results of 512-D NIP descriptors, as the baseline uncompressed scheme.

We observe that NIP+RBMH outperforms other methods at most code sizes on all data sets.
First, there is a significant improvement at smaller code sizes like 32 bits, due to the proposed batch-level regularization:
0.457 vs. 0.369 in terms of mAP, compared to the second performing method RBM on {\it Holidays} at 32 bits.
Second, the improvements of NIP+RBMH over other methods becomes smaller as code size increases (except SKLSH).
For code size larger than 256 bits, the performances of all methods approach the upper bound, i.e., uncompressed NIP descriptors. 
Finally, compared to uncompressed NIP descriptors, 
there is a marginal drop for all methods on {\it UKB} at 256 bits, while performance gap is larger for other datasets.

\begin{figure}[ht] 
    \begin{minipage}[b]{.72\textwidth}
    \centering
      \begin{tabular}{cc}
        \includegraphics[width=.5\textwidth]{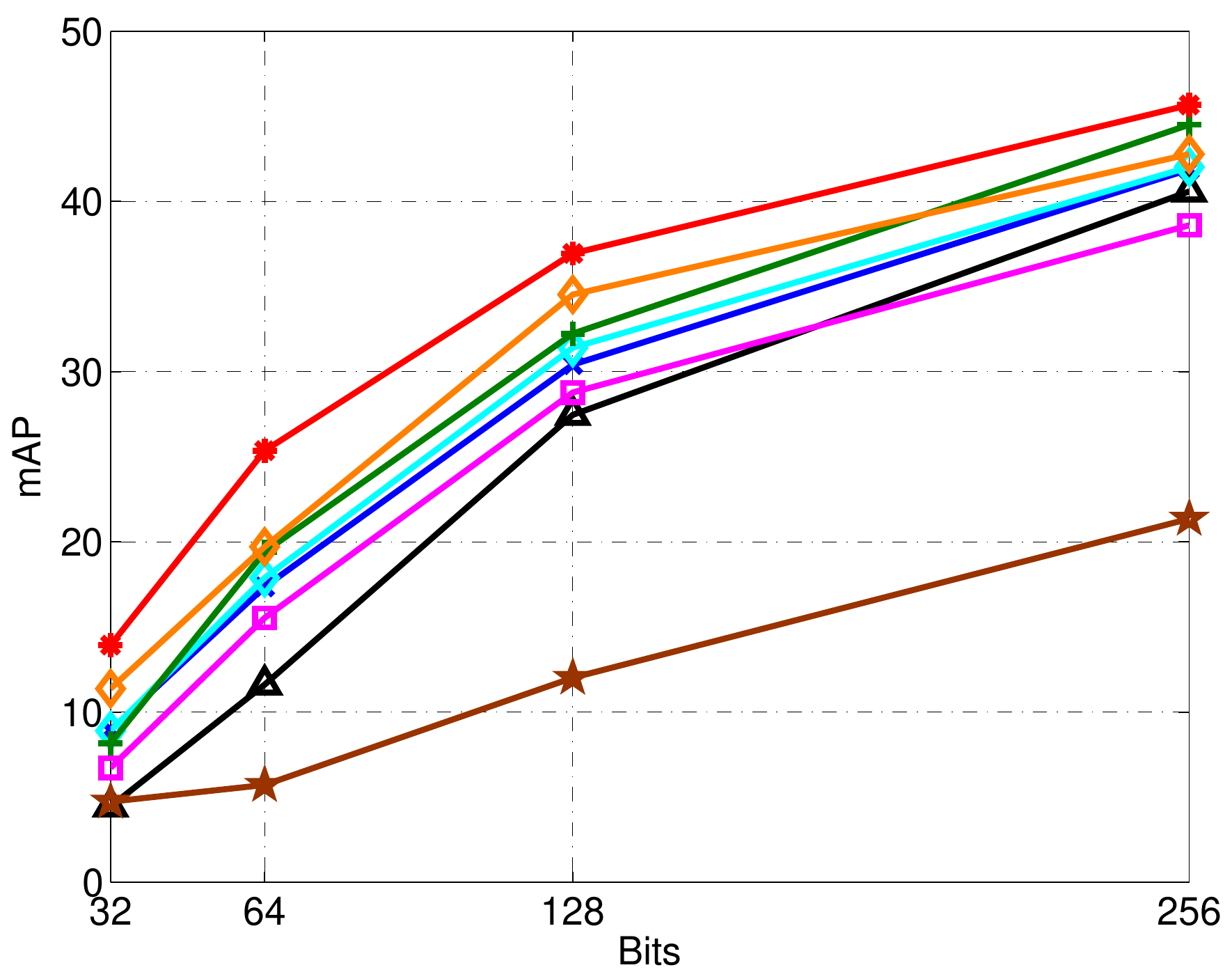} & 
        \includegraphics[width=.5\textwidth]{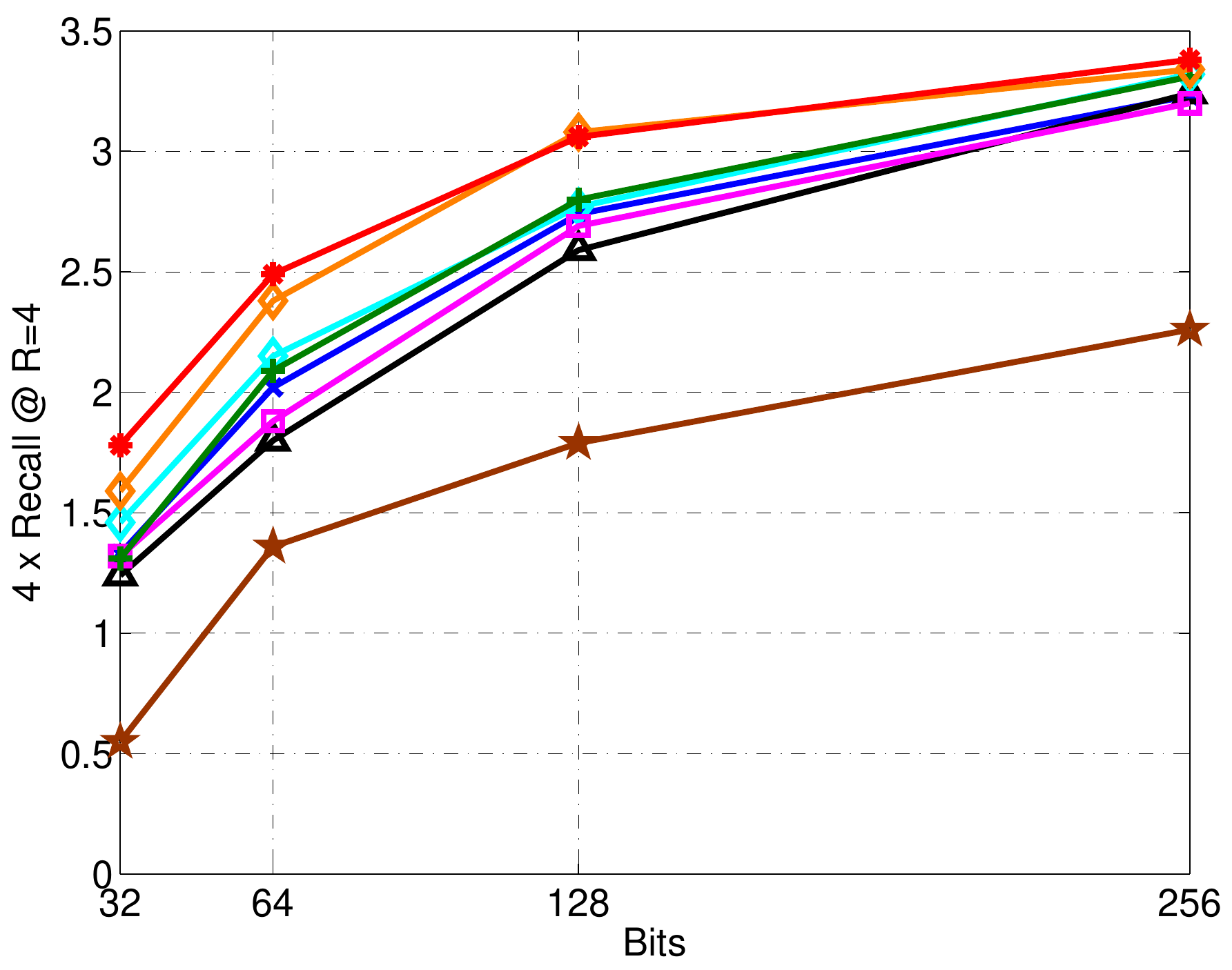} \\ 
        (a) Holidays+1M & (b) UKB+1M \\        
		\includegraphics[width=.5\textwidth]{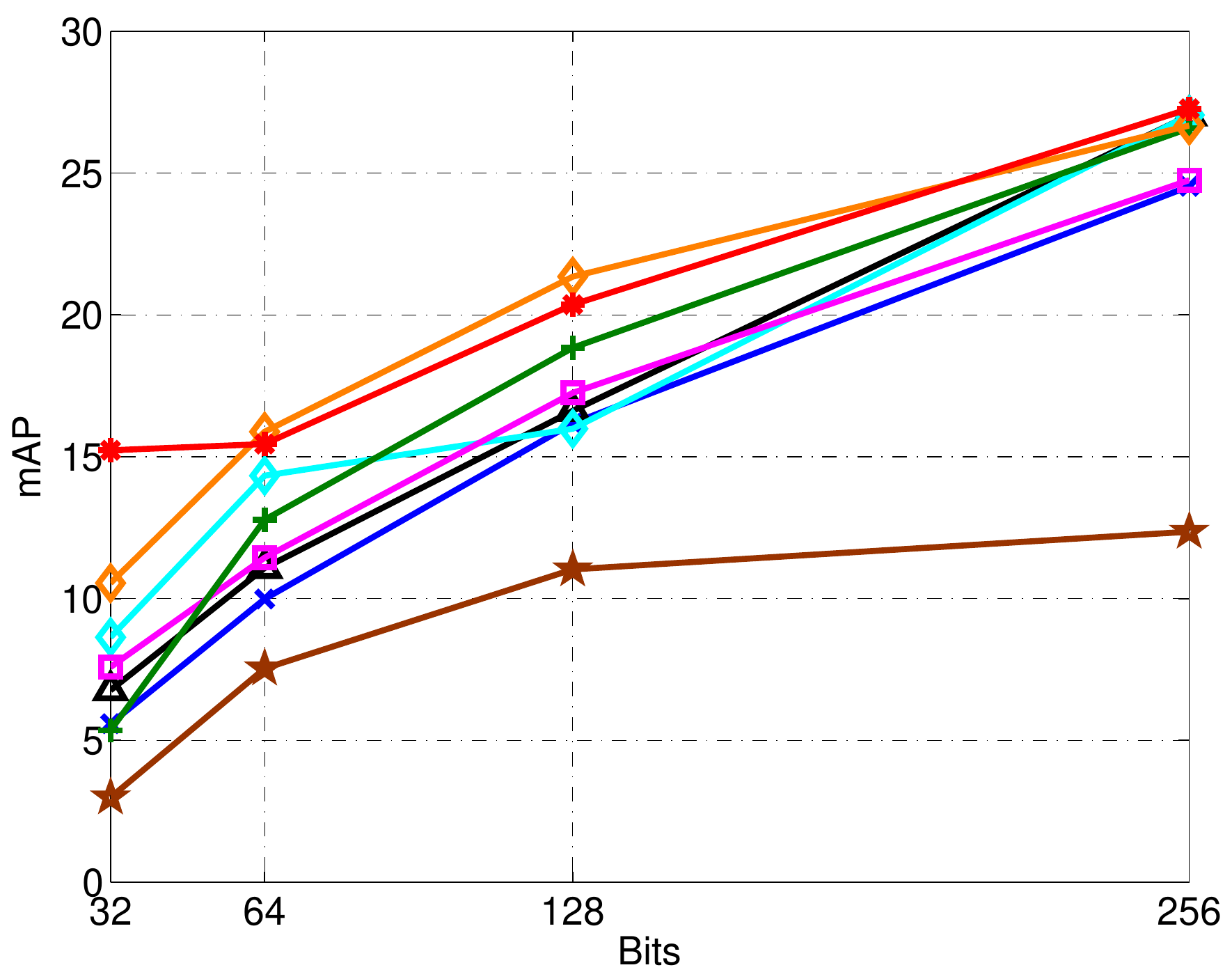} &
		\includegraphics[width=.5\textwidth]{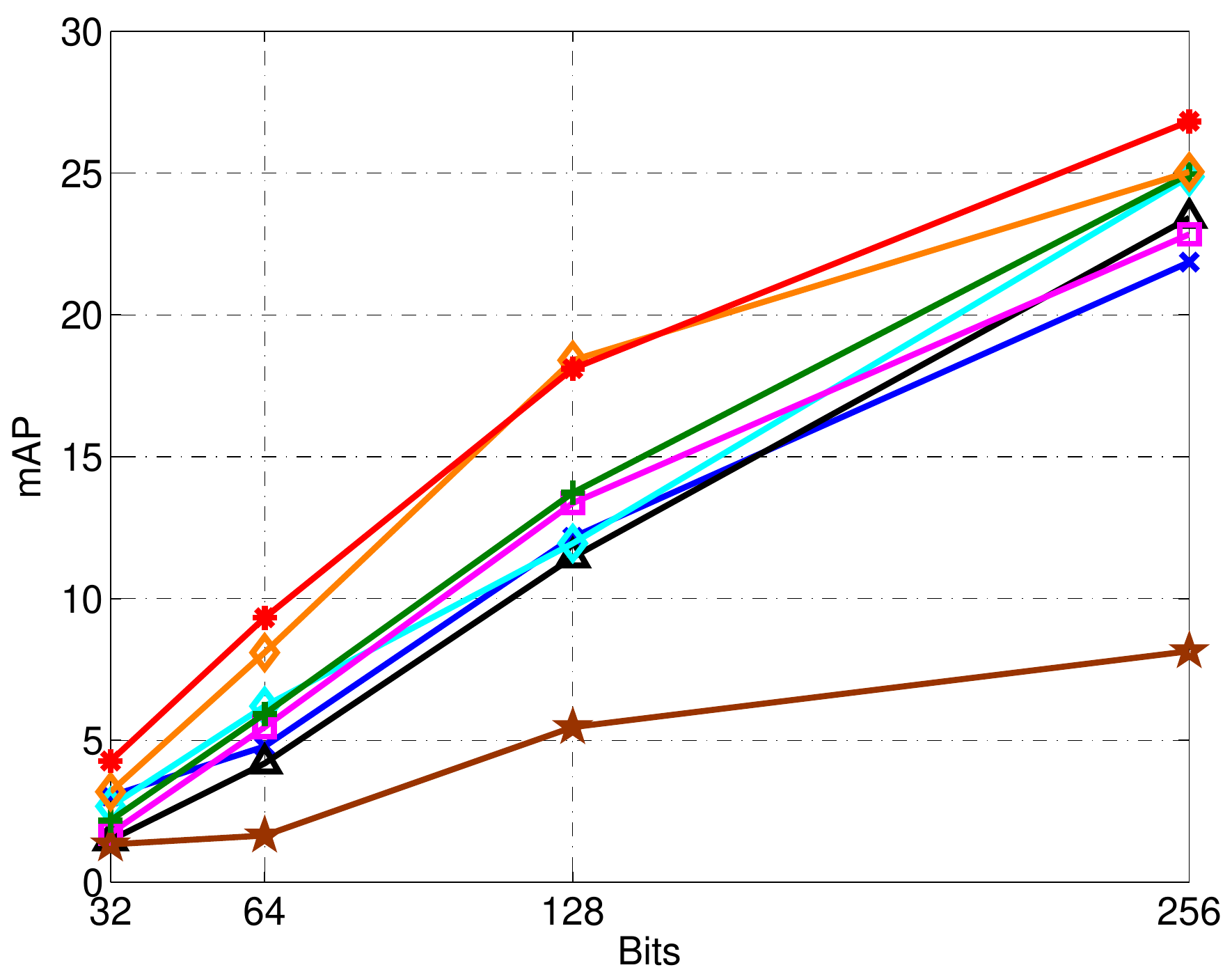} \\
		(c) Oxford5K+1M & (d) Graphics+1M        
      \end{tabular}
    \end{minipage}%
    \begin{minipage}[b]{.2\textwidth}
        \centering
        \includegraphics[width=0.85\textwidth]{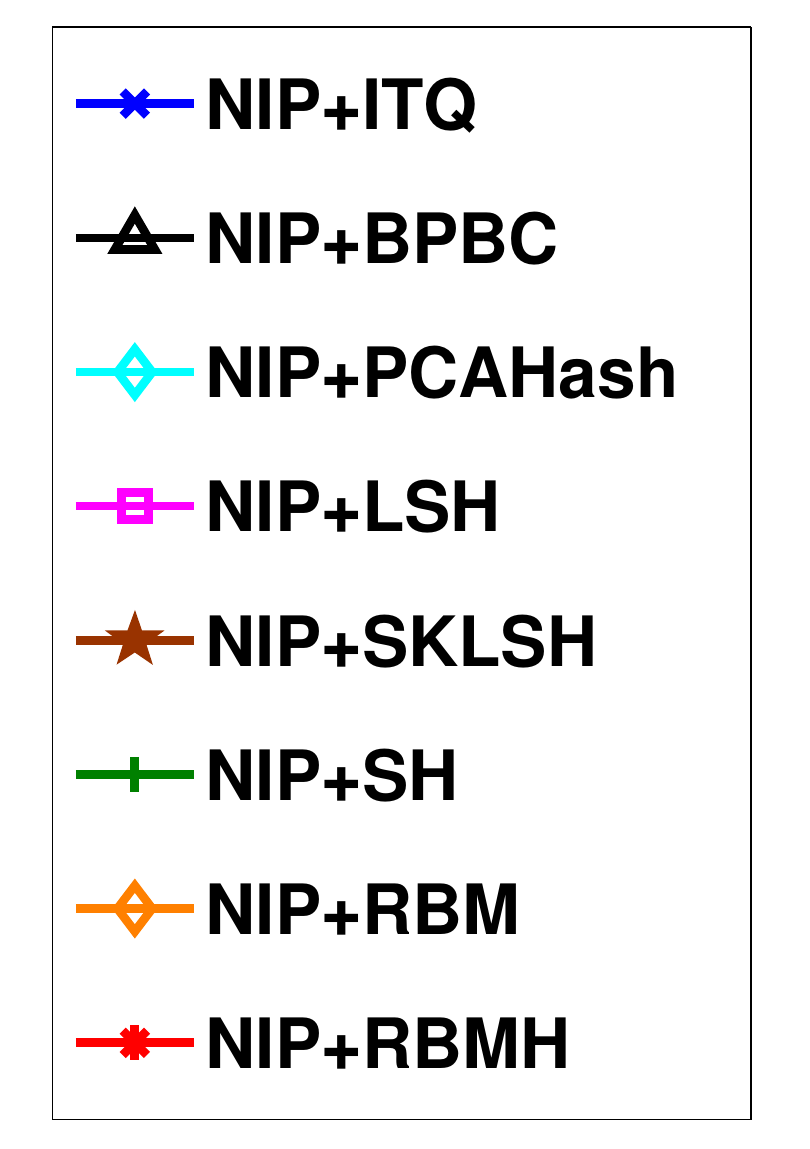}
    \end{minipage}%
\caption{\footnotesize Comparison of RBMH with other hashing methods on large scale retrieval experiments. All methods are based on the best NIP descriptors.}
\label{fig:reshashlarge}
\vspace*{-3mm}
\end{figure}

\textbf{Comparing NIP+RBMH with state-of-the-art} including methods compressing VLAD/FV with direct binarization~\cite{Perronnin_CVPR_10}, 
hashing~\cite{asdembedding} and PQ~\cite{PQFisher,Zhang_2015_CVPR}, methods based on compact deep descriptors~\cite{Yandex,sharif2015baseline}.

As shown in Table~\ref{tab:stateofthearthash},
first, a simple binarization strategy applied to our best performing NIP descriptor is sufficient to 
obtain significantly better accuracy than~\cite{Perronnin_CVPR_10,Yandex} at comparable code size (512 bits), 
e.g., 3.7 vs. 2.79 in~\cite{Perronnin_CVPR_10} for 4$\times$ Recall @ $R=4$ on {\it UKB}.
Second, NIP+RBMH outperforms state-of-the-art by a significant margin at comparable code sizes (from 32 to 256 bits).
NIP+RBMH achieves the best performance on {\it Holidays} at small code size (128 bits), 0.705 vs. 0.644 mAP reported in ~\cite{icml2014c2_zhangd14}
(to our knowledge, the state-of-the-art on this dataset with 128-bit descriptors).
Note that Hamming distance is used for our binary descriptors, while other methods like PQ variants employ Euclidean distances (L2 or ADC),
which typically result in higher accuracy than Hamming distance, at the expense of higher computational cost.

\textbf{Large scale experiments}.
In Figure~\ref{fig:reshashlarge}, we present large scale retrieval results,
combining the 1 million MIR FLICKR distractor images with each data set respectively.
Trends consistent with small scale retrieval results in Figure~\ref{fig:reshashsmall} are observed.

\section{Conclusions}
\label{sec:conclusions}

In this work, we proposed a method to produce global image descriptors from CNNs which are both compact and robust to typical geometric transformations.
The method provides a practical and mathematically proven way for computing invariant object representations with feed-forward neural networks.
To achieve global geometric invariance, we introduce a series of nested pooling layers at intermediate levels of the deep CNN network.
We further introduce a RBM layer with a novel batch-level regularization scheme for generating compact binary descriptors.
Through a thorough evaluation with state-of-the-art, we show that the proposed method outperforms state-of-the-art by a significant margin.

\clearpage

\bibliographystyle{splncs}
\bibliography{egbib,marbib}

\begin{thebibliography}{10}

\bibitem{Perronnin_CVPR_10}
Perronnin, F., Liu, Y., Sanchez, J., Poirier, H.:
\newblock {Large-scale Image Retrieval with Compressed Fisher Vectors}.
\newblock In: Computer Vision and Pattern Recognition (CVPR). (2010)

\bibitem{PQFisher}
J{\'e}gou, H., Perronnin, F., Douze, M., S{\'a}nchez, J., P{\'e}rez, P.,
  Schmid, C.:
\newblock {Aggregating local image descriptors into compact codes}.
\newblock {IEEE Transactions on Pattern Analysis and Machine Intelligence
  (PAMI)} \textbf{34}(9) (2012)  1704--1716

\bibitem{AlexNet}
Krizhevsky, A., Sutskever, I., Hinton, G.E.:
\newblock {Imagenet classification with deep convolutional neural networks}.
\newblock In: Neural Information Processing Systems (NIPS). (2012)

\bibitem{Yandex}
Babenko, A., Slesarev, A., Chigorin, A., Lempitsky, V.:
\newblock {Neural Codes for Image Retrieval}.
\newblock In: European Conference on Computer Vision (ECCV). (2014)

\bibitem{Simonyan2014}
Simonyan, K., Zisserman, A.:
\newblock {Very Deep Convolutional Networks for Large-Scale Image Recognition}.
\newblock In: International Conference on Learning Representations (ICLR).
  (2015)

\bibitem{deepface}
Taigman, Y., Yang, M., Ranzato, M., Wolf, L.:
\newblock {DeepFace: Closing the Gap to Human-Level Performance in Face
  Verification}.
\newblock In: Computer Vision and Pattern Recognition (CVPR). (2014)

\bibitem{deepid}
Sun, Y., Wang, X., Tang, X.:
\newblock Deep learning face representation from predicting 10,000 classes.
\newblock In: Computer Vision and Pattern Recognition (CVPR). (2014)

\bibitem{deeppose}
Toshev, A., Szegedy, C.:
\newblock Deeppose: Human pose estimation via deep neural networks.
\newblock In: Computer Vision and Pattern Recognition (CVPR). (2014)

\bibitem{TEDA}
J{\'e}gou, H., Zisserman, A.:
\newblock Triangulation embedding and democratic aggregation for image search.
\newblock In: Computer Vision and Pattern Recognition (CVPR). (2014)

\bibitem{confused}
Tolias, G., Avrithis, Y., Jegou, H.:
\newblock To aggregate or not to aggregate: Selective match kernels for image
  search.
\newblock In: International Conference on Computer Vision (ICCV). (2013)

\bibitem{REVV1}
Chen, D.M., Tsai, S.S., Chandrasekhar, V., Takacs, G., Vedantham, R.,
  Grzeszczuk, R., Girod, B.:
\newblock {Residual Enhanced Visual Vector as a Compact Signature for Mobile
  Visual Search}.
\newblock In: Signal Processing. (2012)

\bibitem{SFCV}
Lin, J., Duan, L.Y., Huang, T., Gao, W.:
\newblock {Robust Fisher Codes for Large Scale Image Retrieval}.
\newblock In: International Conference on Acoustics and Signal Processing
  (ICASSP). (2013)

\bibitem{Razavian2014}
Razavian, A.S., Azizpour, H., Sullivan, J., Carlsson, S.:
\newblock {CNN Features Off-the-Shelf: An Astounding Baseline for Recognition}.
\newblock In: Computer Vision and Pattern Recognition Workshops. (2014)

\bibitem{NewPaperA}
Gong, Y., Wang, L., Guo, R., Lazebnik, S.:
\newblock Multi-scale orderless pooling of deep convolutional activation
  features.
\newblock In: European Conference on Computer Vision (ECCV). (2014)

\bibitem{sharif2015baseline}
Razavian, A.S., Sullivan, J., Maki, A., Carlsson, S.:
\newblock A baseline for visual instance retrieval with deep convolutional
  networks.
\newblock In: International Conference on Learning Representations (ICLR).
  (2015)

\bibitem{babenko2}
Babenko, A., Lempitsky, V.:
\newblock {Aggregating local deep features for image retrieval}.
\newblock In: {International Conference on Computer Vision (ICCV)}. (2015)

\bibitem{ToliasSJ15}
Tolias, G., Sicre, R., J{\'{e}}gou, H.:
\newblock Particular object retrieval with integral max-pooling of {CNN}
  activations.
\newblock In: arXiv:1511.05879. (2015)

\bibitem{Razavian15}
Azizpour, H., Razavian, A.S., Sullivan, J., Maki, A., Carlsson, S.:
\newblock From generic to specific deep representations for visual recognition.
\newblock In: Computer Vision and Pattern Recognition Workshops. (2015)

\bibitem{girshick2014rcnn}
Girshick, R., Donahue, J., Darrell, T., Malik, J.:
\newblock Rich feature hierarchies for accurate object detection and semantic
  segmentation.
\newblock In: Computer Vision and Pattern Recognition (CVPR). (2014)

\bibitem{LSH}
Datar, M., Immorlica, N., Indyk, P., Mirrokni, V.S.:
\newblock {Locality-Sensitive Hashing Scheme based on p-stable Distributions}.
\newblock In: Annual Symposium on Computational Geometry. (2004)

\bibitem{ITQ}
Gong, Y., Lazebnik, S., Gordo, A., Perronnin, F.:
\newblock {Iterative Quantization: A Procrustean Approach to Learning Binary
  Codes for Large-scale Image Retrieval}.
\newblock In: IEEE Transaction on Pattern Analysis and Machine Intelligence
  (PAMI). (2013)

\bibitem{SpectralHashing}
Weiss, Y., Torralba, A., Fergus, R.:
\newblock {Spectral Hashing}.
\newblock In: Neural Information Processing Systems (NIPS). (2008)

\bibitem{HintonScience}
Hinton, G.E., Salakhutdinov, R.R.:
\newblock {Reducing the dimensionality of data with neural networks}.
\newblock Science \textbf{313} (2006)  504--507

\bibitem{hintonSparsity}
Nair, V., Hinton, G.:
\newblock {3{D} Object Recognition with Deep Belief Nets}.
\newblock In: Neural Information Processing Systems (NIPS). (2009)

\bibitem{icml2014c2_zhangd14}
Zhang, T., Du, C., Wang, J.:
\newblock Composite quantization for approximate nearest neighbor search.
\newblock In: International Conference on Machine Learning (ICML). (2014)

\bibitem{Zhang_2015_CVPR}
Zhang, T., Qi, G.J., Tang, J., Wang, J.:
\newblock Sparse composite quantization.
\newblock In: Computer Vision and Pattern Recognition (CVPR). (2015)

\bibitem{itheory1}
Anselmi, F., Leibo, J.Z., Rosasco, L., Mutch, J., Tacchetti, A., Poggio, T.:
\newblock Unsupervised learning of invariant representations in hierarchical
  architectures.
\newblock arXiv:1311.4158 (2013)

\bibitem{itheory2}
Anselmi, F., Poggio, T.:
\newblock Representation learning in sensory cortex: a theory.
\newblock CBMM memo n 26 (2010)

\bibitem{itheory3}
Anselmi, F., Leibo, J.Z., Rosasco, L., Mutch, J., Tacchetti, A., Poggio, T.:
\newblock Magic materials: a theory of deep hierarchical architectures for
  learning sensory representations.
\newblock CBCL paper (2013)

\bibitem{poggioface}
Liao, Q.L., Leibo, J.Z., Poggio, T.:
\newblock Learning invariant representations and applications to face
  verification.
\newblock In: Neural Information Processing Systems (NIPS). (2013)

\bibitem{poggiomusic}
Zhang, C., Evangelopoulos, G., Voinea, S., Rosasco, L., Poggio, T.:
\newblock A deep representation for invariance and music classification.
\newblock In: International Conference on Acoustics and Signal Processing
  (ICASSP). (2014)

\bibitem{hintonCD}
Hinton, G.E.:
\newblock Training products of experts by minimizing contrastive divergence.
\newblock Neural Computation \textbf{14}(8) (2002)  1771{\textendash}1800

\bibitem{hanlinSparsity}
Goh, H., Thome, N., Cord, M., Lim, J.H.:
\newblock Unsupervised and supervised visual codes with restricted {B}oltzmann
  machines.
\newblock In: European Conference on Computer Vision (ECCV). (2012)

\bibitem{Jegou08}
J\'{e}gou, H., Douze, M., Schmid, C.:
\newblock {Hamming Embedding and Weak Geometric Consistency for Large Scale
  Image Search}.
\newblock In: European Conference on Computer Vision (ECCV). (2008)

\bibitem{Nister06}
Nist\'er, D., Stew\'enius, H.:
\newblock {Scalable Recognition with a Vocabulary Tree}.
\newblock In: Computer Vision and Pattern Recognition (CVPR). (2006)

\bibitem{Philbin07}
Philbin, J., Chum, O., Isard, M., Sivic, J., Zisserman, A.:
\newblock {Object Retrieval with Large Vocabularies and Fast Spatial Matching}.
\newblock In: Computer Vision and Pattern Recognition (CVPR). (2007)

\bibitem{SVMSDataSet}
Chandrasekhar, V., D.M.Chen, S.S.Tsai, N.M.Cheung, H.Chen, G.Takacs, Y.Reznik,
  R.Vedantham, R.Grzeszczuk, J.Back, B.Girod:
\newblock {Stanford Mobile Visual Search Data Set}.
\newblock In: {ACM Multimedia Systems Conference (MMSys)}. (2011)

\bibitem{MPEGDataset2}
ISO/IEC-JTC1/SC29/WG11/N12202:
\newblock Evaluation Framework for Compact Descriptors for Visual Search.
  (2011)

\bibitem{mirflickr}
Mark J.~Huiskes, B.T., Lew, M.S.:
\newblock {New Trends and Ideas in Visual Concept Detection: The MIR Flickr
  Retrieval Evaluation Initiative}.
\newblock In: ACM International Conference on Multimedia Information Retrieval.
  (2010)

\bibitem{DengImagenet}
Deng, J., Dong, W., Socher, R., Li, L.J., Li, K., Fei-Fei, L.:
\newblock Imagenet: A large-scale hierarchical image database.
\newblock In: Computer Vision and Pattern Recognition (CVPR). (2009)

\bibitem{FisherColor}
Gordoa, A., Rodriguez-Serrano, J., Perronnin, F., Valveny, E.:
\newblock Leveraging category-level labels for instance-level image retrieval.
\newblock In: Computer Vision and Pattern Recognition (CVPR). (2012)

\bibitem{Perronnin_2015_CVPR}
Perronnin, F., Larlus, D.:
\newblock Fisher vectors meet neural networks: A hybrid classification
  architecture.
\newblock In: Computer Vision and Pattern Recognition (CVPR). (2015)

\bibitem{girshick15fastrcnn}
Girshick, R.:
\newblock Fast {R-CNN}.
\newblock In: International Conference on Computer Vision ({ICCV}). (2015)

\bibitem{asdembedding}
Gordo, A., Perronnin, F., Gong, Y., Lazebnik, S.:
\newblock Asymmetric distances for binary embeddings.
\newblock IEEE Transactions on Pattern Analysis and Machine Intelligence (PAMI)
  \textbf{36}(1) (2014)  33--47

\bibitem{BPBC}
Gong, Y., Kumar, S., Rowley, H., Lazebnik, S.:
\newblock {Learning Binary Codes for High-Dimensional Data Using Bilinear
  Projections.}
\newblock In: Computer Vision and Pattern Recognition (CVPR). (2013)

\bibitem{SKLSH}
Raginsky, M., Lazebnik, S.:
\newblock {Locality-Sensitive Binary Codes from Shift-Invariant Kernels}.
\newblock In: Neural Information Processing Systems (NIPS). (2009)

\end{thebibliography}
\end{document}